%% file: main.tex
\documentclass[11pt,a4paper]{extarticle}
\usepackage{styles/technical_report}

\input{styles/iclr2027/math_commands.tex}

\usepackage{hyperref}
\usepackage{url}
\definecolor{reportReference}{HTML}{70A3F3}
\hypersetup{
    colorlinks=true,
    linkcolor=reportElectricBlue,
    citecolor=reportReference,
    urlcolor=reportReference,
    filecolor=reportReference,
    pdfborder={0 0 0}
}
\usepackage{graphicx}
\usepackage{float}
\usepackage{assets/vendor/wrapfig}
\usepackage{algorithm}
\usepackage{algpseudocode}
\usepackage{tikz}
\usetikzlibrary{calc}
\usepackage{booktabs}
\usepackage{multirow}
\usepackage{colortbl}
\usepackage{array}

\input{assets/metadata.tex}
\input{assets/colors.tex}
\input{assets/prompt_blocks.tex}

\input{assets/algorithm_blocks.tex}
\input{assets/project_commands.tex}
\input{assets/notation.tex}
\input{assets/typography/correspondence-font.tex}
\input{assets/project_links.tex}

\title{\papertitle}
\author{%
\ReportAuthor{Shaohang Wei}{1,\textdaggerdbl,*}, \ReportAuthor{Feifan Song}{1}, \ReportAuthor{Guangyue Peng}{1}, \ReportAuthor{Wenhao Yu}{3}, \ReportAuthor{Wei Li}{1},\\
\ReportAuthor{Wen Luo}{1}, \ReportAuthor{Yang Xu}{4}, \ReportAuthor{Yufan Shen}{2}, \ReportAuthor{Luke Mao}{2}, \ReportAuthor{Yang Du}{2},\\
\ReportAuthor{Asher Qin}{2}, \ReportAuthor{Houfeng Wang}{1,\textdagger}}
\newcommand{\ReportAffiliations}{%
\mbox{\textsuperscript{1}Peking University}\hspace{1.8em}%
\mbox{\textsuperscript{2}Tencent}\hspace{1.8em}%
\mbox{\textsuperscript{3}CUHK}\hspace{1.8em}%
\mbox{\textsuperscript{4}Nanjing University}}
\newcommand{\ReportAuthorNotes}{%
\mbox{\textsuperscript{\textdaggerdbl}Project leader}\hspace{22.5pt}%
\mbox{\textsuperscript{\textdagger}Corresponding authors}}

\begin{document}

\maketitle

\let\ReportDefaultFootnoteRule\footnoterule
\renewcommand{\footnoterule}{\kern-3pt\hrule width12pc height0.4pt\kern2.6pt}
\begingroup
\renewcommand{\thefootnote}{\fnsymbol{footnote}}
\footnotetext[1]{%
\mbox{%
Work done during the internship at Tencent.
Correspondence to \ReportCorrespondenceEmail{shaohang@stu.pku.edu.cn}.
}%
}
\endgroup

\addtocontents{toc}{\protect\setcounter{tocdepth}{-1}}

\input{contents/00_abstract.tex}
\input{figs/main_figure/figure1.tex}
\clearpage
\let\footnoterule\ReportDefaultFootnoteRule
\input{contents/01_introduction.tex}
\input{contents/02_related_work.tex}

\input{contents/03_problem_setup.tex}
\input{contents/04_method.tex}
\input{contents/05_experiments.tex}
\input{contents/06_analysis_and_discussion.tex}

\input{contents/08_conclusion.tex}

\clearpage
\begingroup
\addtolength{\bibsep}{-0.8pt}
\bibliography{references}
\endgroup
\bibliographystyle{styles/iclr2027/report}

\appendix
\addtocontents{toc}{\protect\setcounter{tocdepth}{2}}
\input{appendix/contents.tex}
\input{appendix/a_extended_method.tex}
\input{appendix/b_experimental_details.tex}

\input{appendix/c_additional_results.tex}

\input{appendix/d_proofs.tex}

\input{appendix/case_study.tex}

\input{appendix/benchmarks.tex}

\end{document}

%% file: styles/iclr2027/math_commands.tex
\usepackage{amsmath,amsfonts,bm}

\def\eqref#1{equation~\ref{#1}}
\def\Eqref#1{Equation~\ref{#1}}

\def\1{\bm{1}}

\DeclareMathAlphabet{\mathsfit}{\encodingdefault}{\sfdefault}{m}{sl}
\SetMathAlphabet{\mathsfit}{bold}{\encodingdefault}{\sfdefault}{bx}{n}

\newcommand{\KL}{D_{\mathrm{KL}}}

%% file: assets/metadata.tex
\newcommand{\papertitle}{On-Policy Visual Evidence Distillation}

\newcommand{\ReportWebsiteURL}{https://sylvain-wei.github.io/ReVuE/}
\newcommand{\ReportCodeURL}{https://github.com/sylvain-wei/ReVuE}

%% file: assets/colors.tex
\definecolor{oiBlack}{HTML}{000000}
\definecolor{oiOrange}{HTML}{E69F00}
\definecolor{oiSkyBlue}{HTML}{56B4E9}
\definecolor{oiBluishGreen}{HTML}{009E73}
\definecolor{oiYellow}{HTML}{F0E442}
\definecolor{oiBlue}{HTML}{0072B2}
\definecolor{oiVermillion}{HTML}{D55E00}
\definecolor{oiReddishPurple}{HTML}{CC79A7}

\definecolor{tableModelGray}{gray}{0.92}
\definecolor{tableOursBlue}{HTML}{CFEFFF}
\definecolor{tableColdGray}{HTML}{808080}
\definecolor{tableExpertGray}{HTML}{595959}

\definecolor{algorithmSamplingBlue}{HTML}{F1F6FC}
\definecolor{algorithmReflectionBlue}{HTML}{DFEBF8}
\definecolor{algorithmUpdateYellow}{HTML}{FFF8E5}
\definecolor{algorithmBlueNote}{HTML}{4E7094}
\definecolor{algorithmYellowNote}{HTML}{8D7335}

\definecolor{algorithmVerifyRed}{HTML}{B43C3C}
\definecolor{algorithmStatesGreen}{HTML}{2E7D50}
\definecolor{algorithmAssembleBlue}{HTML}{3569A5}

\definecolor{tokenImpactBlue}{HTML}{174EA6}
\definecolor{tokenImpactOrange}{HTML}{C45100}

\definecolor{groupWeightBlue}{HTML}{4285F4}
\definecolor{groupWeightGray}{HTML}{7A7A7A}

%% file: assets/prompt_blocks.tex
\usepackage[breakable,skins]{tcolorbox}
\colorlet{promptFrame}{reportGoogleBlue}
\colorlet{promptBackground}{reportPaleBlue}
\newtcolorbox{PromptBox}[1]{
    enhanced,breakable,
    title={#1},title after break={#1 (continued)},
    fonttitle=\normalfont\rmfamily\fontsize{9}{10}\selectfont,
    fontupper=\normalfont\rmfamily\fontsize{9}{10}\selectfont,
    colback=promptBackground,colframe=promptFrame,coltitle=white,
    boxrule=0.5mm,arc=1mm,outer arc=1.5mm,
    boxsep=1mm,left=1mm,right=1mm,top=1mm,bottom=1mm,
    toptitle=0mm,bottomtitle=0mm,
    before skip=9pt,after skip=9pt,
    pad at break*=1mm,
    before upper={%
    \setlength{\parindent}{0pt}%
    \setlength{\parskip}{10pt}%
    \hyphenpenalty=10000\relax
    }
}

%% file: assets/algorithm_blocks.tex
\newcommand{\AlgorithmBlockMark}[1]{%
    \tikz[remember picture,overlay]{\coordinate (#1) at (0,0);}%
}
\newcommand{\AlgorithmBlockOrigin}{%
    \tikz[remember picture,overlay]{%
        \coordinate (revue-alg-left) at (-\dimexpr\labelwidth+\labelsep\relax,0);
        \coordinate (revue-alg-right) at (\linewidth,0);
    }%
}

\newcommand{\RevueAlgorithmBlocks}{%
    \AddToHook{shipout/background}[revue-algorithm-blocks]{%
        \ifnum\value{page}=\getpagerefnumber{alg:revue-training}\relax
            \begin{tikzpicture}[remember picture,overlay]
                \fill[algorithmSamplingBlue,rounded corners=1.5pt]
                    ($(revue-alg-left |- revue-sampling-start)+(0,8pt)$)
                    rectangle
                    ($(revue-alg-right |- revue-sampling-end)+(0,-3pt)$);
                \fill[algorithmReflectionBlue,rounded corners=1pt]
                    ($(revue-alg-left |- revue-reflection-start)+(2pt,8pt)$)
                    rectangle
                    ($(revue-alg-right |- revue-reflection-end)+(-2pt,-3pt)$);
                \fill[algorithmUpdateYellow,rounded corners=1.5pt]
                    ($(revue-alg-left |- revue-update-start)+(0,7pt)$)
                    rectangle
                    ($(revue-alg-right |- revue-update-end)+(0,-3pt)$);

                \draw[algorithmBlueNote!65,thin]
                    ($(revue-alg-right |- revue-sampling-start)+(-95pt,8pt)$)
                    -- ++(4pt,0)
                    -- ($(revue-alg-right |- revue-sampling-end)+(-91pt,-3pt)$)
                    -- ++(-4pt,0);
                \draw[algorithmBlueNote!75,thin]
                    ($(revue-alg-right |- revue-reflection-start)+(-85pt,8pt)$)
                    -- ++(4pt,0)
                    -- ($(revue-alg-right |- revue-reflection-end)+(-81pt,-3pt)$)
                    -- ++(-4pt,0);
                \draw[algorithmYellowNote!65,thin]
                    ($(revue-alg-right |- revue-update-start)+(-95pt,7pt)$)
                    -- ++(4pt,0)
                    -- ($(revue-alg-right |- revue-update-end)+(-91pt,-3pt)$)
                    -- ++(-4pt,0);

                \coordinate (revue-sampling-note) at
                    ($(revue-sampling-start)!0.5!(revue-rollout)$);
                \coordinate (revue-reflection-note) at
                    ($(revue-reflection-start)!0.5!(revue-reflection-end)$);
                \coordinate (revue-update-note) at
                    ($(revue-update-start)!0.5!(revue-update-end)$);
                \tikzset{algorithm note/.style={anchor=west,align=left,
                    inner sep=0pt,text width=77pt,
                    font=\fontsize{8}{9.4}\selectfont\itshape}}
                \node[algorithm note,text=algorithmBlueNote]
                    at ($(revue-alg-right |- revue-sampling-note)+(-84pt,2.5pt)$)
                    {Sampling from\\student};
                \node[algorithm note,text=algorithmBlueNote,text width=66pt]
                    at ($(revue-alg-right |- revue-reflection-note)+(-74pt,2.5pt)$)
                    {Reflection via\\critic};
                \node[algorithm note,text=algorithmYellowNote]
                    at ($(revue-alg-right |- revue-update-note)+(-84pt,2.5pt)$)
                    {Teacher evaluation\\Reweighting\\Gradient update};
            \end{tikzpicture}%
        \fi
    }%
}

%% file: assets/project_commands.tex
\newcommand{\method}{\textsc{ReVuE}}
\newcommand{\ours}{\method{}(Ours)}

\newcommand{\singleRuleTableGap}{\noalign{\vskip 2pt}}

\newcommand{\mainTableModelHeader}[1]{%
  \rule[-3.5pt]{0pt}{14pt}%
  \raisebox{\dimexpr3.5pt-(\height-\depth)/2\relax}{%
    \fontsize{8.5}{10}\selectfont\bfseries #1}%
}

%% file: assets/notation.tex
\newtheorem{lemma}{Lemma}
\renewcommand{\KL}{\mathbb{D}^{\mathrm{KL}}}

%% file: assets/typography/correspondence-font.tex
\pdfmapline{+reportcmtt0800 SFTT0800 "T1Encoding ReEncodeFont" <assets/typography/cm-super-t1.enc <assets/texmf/fonts/type1/public/cm-super/sftt0800.pfb}
\DeclareFontFamily{T1}{reportcmtt}{\hyphenchar\font=-1}
\DeclareFontShape{T1}{reportcmtt}{m}{n}{<->reportcmtt0800}{}
\newcommand{\ReportCorrespondenceEmail}[1]{%
  {\fontencoding{T1}\fontfamily{reportcmtt}\fontseries{m}\fontshape{n}\selectfont #1}}

%% file: assets/project_links.tex
\pdfmapline{+fvmr8r BeraSansMono-Roman "TeXBase1Encoding ReEncodeFont" <8r.enc <assets/texmf/fonts/type1/public/bera/fvmr8a.pfb}
\pdfmapline{+fa5free1solid FontAwesome5Free-Solid "fa5free1 ReEncodeFont" <[assets/typography/project-links/fa5free1.enc <assets/texmf/fonts/type1/public/fontawesome5/FontAwesome5Free-Solid.pfb}
\pdfmapline{+fa5brands0 FontAwesome5Brands-Regular "fa5brands0 ReEncodeFont" <[assets/typography/project-links/fa5brands0.enc <assets/texmf/fonts/type1/public/fontawesome5/FontAwesome5Brands-Regular.pfb}
\DeclareFontFamily{T1}{reportlinkmono}{\hyphenchar\font=-1}
\DeclareFontShape{T1}{reportlinkmono}{m}{n}{<->fvmr8t}{}
\DeclareFontFamily{U}{reportlinkglobe}{}
\DeclareFontShape{U}{reportlinkglobe}{m}{n}{<->fa5free1solid}{}
\DeclareFontFamily{U}{reportlinkgithub}{}
\DeclareFontShape{U}{reportlinkgithub}{m}{n}{<->fa5brands0}{}
\definecolor{reportLinkBackground}{HTML}{F8F8F8}
\definecolor{reportLinkText}{HTML}{8C8E90}
\newcommand{\ReportGlobeIcon}{{\fontencoding{U}\fontfamily{reportlinkglobe}\selectfont\char128}}
\newcommand{\ReportGithubIcon}{{\fontencoding{U}\fontfamily{reportlinkgithub}\selectfont\char167}}
\newcommand{\ReportLinkButton}[3]{%
  \href{#1}{\tcbox[on line,
    colback=reportLinkBackground,colframe=reportLinkBackground,
    coltext=reportLinkText,boxrule=.4pt,arc=4pt,boxsep=2pt,
    left=4pt,right=4pt,top=2pt,bottom=2pt]{%
      \fontencoding{T1}\fontfamily{reportlinkmono}\fontseries{m}\fontshape{n}%
      \fontsize{9}{11}\selectfont #2\enspace #3}}}
\newcommand{\ReportProjectLinks}{%
  \begingroup\microtypesetup{expansion=false,protrusion=false}%
  \ReportLinkButton{\ReportWebsiteURL}{\ReportGlobeIcon}{Website}%
  \hspace{2.5pt}%
  \ReportLinkButton{\ReportCodeURL}{\ReportGithubIcon}{Code}%
  \endgroup}

%% file: contents/00_abstract.tex
\begin{abstract}
Visual agents solve problems by interleaving reasoning with image operations, and on-policy distillation (OPD) provides guidance from a strong teacher on student-generated interaction trajectories.
However, image operations change the evidence available for subsequent reasoning, so local errors in evidence acquisition (\textit{Acquire}), reading (\textit{Read}), or answer grounding (\textit{Ground}) can propagate through the trajectory and lead to incorrect answers.
Existing multimodal OPD methods primarily construct or contrast auxiliary views of the original image to strengthen supervision, without explicitly modeling the connections between student actions, resulting observations, and subsequent reasoning.
This limits their ability to provide corrections tailored to different failure stages.
We introduce \underline{\textbf{Re}}flection on \underline{\textbf{V}}is\underline{\textbf{u}}al \underline{\textbf{E}}vidence (\textbf{\method{}}), an on-policy distillation method for visual agents.
\textbf{\method{}} compares multiple student-generated trajectories for the same query, summarizes the observed visual evidence, and diagnoses the first failure across the \textit{Acquire}, \textit{Read}, and \textit{Ground} stages.
The resulting reflections provide training-time context for the teacher.
We group and reweight token-level distillation losses according to how strongly these reflections affect the teacher's predictions.
This design translates trajectory-level evidence diagnosis into targeted token-level supervision, guiding students to improve their visual evidence acquisition and reasoning.
Across 11 benchmarks spanning the Qwen2.5-VL and InternVL3.5 model families, \textbf{\method{}} outperforms all evaluated OPD baselines in weighted-average scores for perception, mathematical reasoning, and general tasks.
\textbf{\method{}} also reduces redundancy in reasoning and tool calls while improving tool-call accuracy and task accuracy.
\end{abstract}

%% file: figs/main_figure/figure1.tex
\begingroup
\setlength{\intextsep}{2pt}
\setlength{\abovecaptionskip}{4pt}
\begin{figure}[H]
  \centering
  \includegraphics[width=\textwidth]{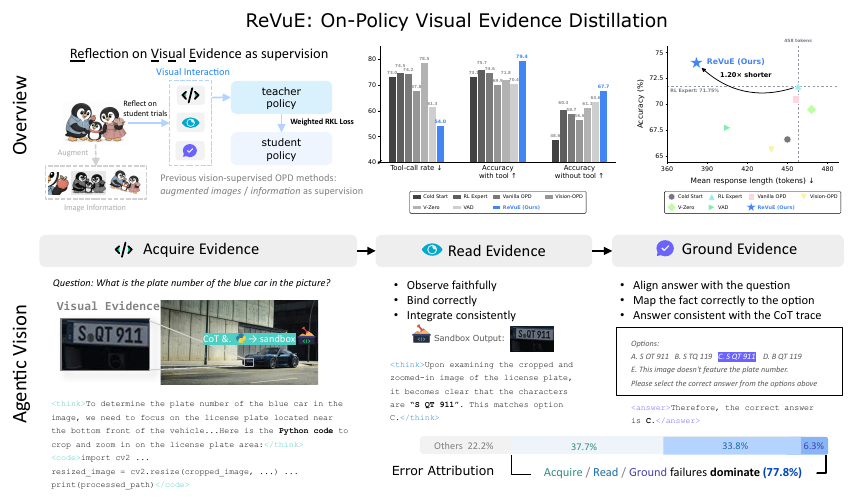}
  \caption{
  \textbf{\method{} overview \textnormal{(with HRBench-8K results)} and an agentic vision example.}
  }
  \label{fig:overview}
\end{figure}
\endgroup

%% file: contents/01_introduction.tex
\section{Introduction}
\label{sec:introduction}

High-resolution image understanding, chart analysis, and document question answering often require models to actively locate task-relevant regions, inspect fine-grained details, and incorporate their observations into subsequent reasoning~\citep{wu2024vstar,qi2025cogcom}.
Visual agents address these demands by interleaving reasoning with image operations, actively gathering information needed to answer a query~\citep{zheng2026deepeyes,zhang2026thyme}.
As strong models develop these capabilities, effectively transferring their expertise to students becomes an important training objective.
On-policy distillation (OPD) provides dense teacher feedback on student-generated trajectories~\citep{agarwal2024onpolicydistillation,gu2024minillm,wei2026verifier}, supervising the actions and reasoning decisions students actually make.
Therefore, constructing teacher supervision tailored to visual interaction is a key step toward translating expert capabilities into reliable student behavior.

Visual interactions introduce a particular challenge: image operations change the evidence available for subsequent reasoning~\citep{qi2025cogcom,zhang2026thyme}. 
Selecting the wrong region can leave critical evidence unavailable~\citep{hou2026codev}; 
obtaining a useful observation does not guarantee that the student reads it correctly; 
and a correctly read fact may still be mapped to the wrong answer. 
We refer to these stages as evidence acquisition (\emph{Acquire}), visual reading (\emph{Read}), and answer grounding (\emph{Ground}). 
Errors at any stage can propagate through later actions and reasoning, ultimately producing an incorrect answer. 
Effective distillation therefore requires identifying the first failed stage in the evidence chain and directing corrections to where the error begins to propagate.

Representative multimodal OPD methods strengthen visual supervision through privileged visual cues or contrasts between teacher predictions under different visual conditions~\citep{tian2026vicur,sun2026vzero,liu2026vaopd}.
Vision-OPD~\citep{yuan2026visionopd} uses evidence-centered crops to guide the teacher, while VAD~\citep{zhang2026vad} estimates the visual contribution to teacher corrections by removing relevant evidence. 
These methods improve how visual information informs supervision, but do not explicitly trace the links between students' executed actions, returned observations, and subsequent reasoning. 
This limits their ability to identify which visual behaviors to reinforce and where the evidence chain first breaks down.
This raises a central question: \emph{how can we turn students' own interaction trajectories into more targeted teacher guidance?}

Multiple attempts at the same query offer a concrete basis for such guidance.
Successful attempts may complement each other: one finds more focused evidence, while another reads the relevant visual facts more accurately.
Failed attempts help identify the first error in \emph{Acquire}, \emph{Read}, or \emph{Ground}.
We introduce \textbf{\method{}} (\underline{\textbf{Re}}flection on \underline{\textbf{V}}is\underline{\textbf{u}}al \underline{\textbf{E}}vidence) to turn these experiences into visual-evidence reflections.
By comparing these trajectories, \method{} selects a compact set of observed images sufficient to establish the relevant visual facts.
It also summarizes how to read the evidence correctly and map the resulting facts to the answer.
The resulting reflections provide training-time context for a strong teacher, helping it reinforce effective behaviors already demonstrated by the student and correct errors at the first failed stage.

Visual-evidence reflection changes the teacher's assessment of student choices, with effects that vary across token positions.
By comparing the teacher's predictions on the same student trajectory with and without reflection, we identify where reflection most strongly changes the teacher's support for student choices.
In early training, we find that these changes concentrate on a small fraction of tokens.
To emphasize supervision at these positions, we group tokens by the magnitude of these changes and apply grouped reweighting to give high-impact positions greater weight in the distillation loss.
Our ablations show that this reweighting improves perception and math performance.
Reflection changes the teacher's predictions and guides how token-level losses are weighted.

We evaluate \method{} on 11 benchmarks across the Qwen2.5-VL~\citep{bai2025qwen25vl} and InternVL3.5~\citep{wang2025internvl35} model families. 
\method{} consistently outperforms all OPD baselines in performance across perception, mathematical reasoning, and general tasks.
It also reduces redundant reasoning and unnecessary tool calls, while improving both tool-call accuracy and overall task accuracy.
On Qwen2.5-VL-7B, masking high-impact positions in the distillation loss reduces performance more than masking an equal number of low-impact positions.
Our critic's stage diagnoses agree closely with human judgments.
Together, these results support the effectiveness of \method{} and the reliability of its stage diagnoses.

\begin{ContributionBox}
\textbf{Our contributions are threefold:}
\begin{itemize}
\item \textbf{Visual evidence reflection across student trajectories.}
We compare multiple successful student attempts on the same query to identify minimal sufficient evidence and summarize successful reading and grounding strategies.
We also locate the first failure among \emph{Acquire}, \emph{Read}, and \emph{Ground}, providing training-time guidance to a strong teacher.
\item \textbf{Reflection-aware grouped reweighting for distillation.} We partition tokens according to how strongly reflection changes the teacher’s predictions, and use group-wise normalization and reweighting to focus distillation updates.
\item \textbf{Cross-model gains and validation of high-impact supervision.}
Across two model families and 11 benchmarks, \method{} improves task and tool-call accuracy over OPD baselines while reducing reasoning and tool-call redundancy.
On Qwen2.5-VL-7B, masking high-impact positions in the loss reduces performance more than masking equally many low-impact positions, supporting greater weight on high-impact supervision.
\end{itemize}
\end{ContributionBox}

%% file: contents/02_related_work.tex
\section{Related Work}
\label{sec:related-work}

\paragraph{Agentic Visual Reasoning.}
Agentic visual reasoning interleaves reasoning with image operations and the resulting observations.
This capability appears in general-purpose models such as o3/o4-mini~\citep{openai2025thinkingimages}, Qwen3-VL~\citep{bai2025qwen3vl}, and Kimi K2.5~\citep{kimiteam2026k25}.
Earlier systems use guided visual search or visual sketches to acquire and organize task-relevant evidence~\citep{wu2024vstar,hu2024visualsketchpad}.
CogCoM~\citep{qi2025cogcom} and LATTE~\citep{ma2025latte} learn sequences of visual operations through trajectory supervision.
DeepEyes~\citep{zheng2026deepeyes}, VTool-R1~\citep{wu2026vtoolr1}, and Pixel Reasoner~\citep{su2025pixelreasoner} use reinforcement learning to improve visual tool use and exploration.
Mini-o3~\citep{lai2026minio3} scales visual search to longer interactions and more diverse reasoning patterns.
Thyme~\citep{zhang2026thyme} and DeepEyesV2~\citep{hong2026deepeyesv2} support broader reasoning through executable code, with DeepEyesV2 also incorporating web search.
CodeV~\citep{hou2026codev} uses process rewards on visual tool inputs and outputs to encourage evidence-consistent tool use.
We study how students' actual visual interactions can inform teacher guidance for on-policy distillation.

\paragraph{On-Policy Distillation.}

On-policy distillation (OPD) supervises student-generated trajectories with teacher distributions, reducing the training--inference mismatch of offline distillation~\citep{agarwal2024onpolicydistillation,gu2024minillm}.
Qwen3~\citep{yang2025qwen3} and Qwen3-VL~\citep{bai2025qwen3vl} also use this approach to train smaller models.
Privileged-context methods condition teachers on reference solutions, additional context, or peer trajectories~\citep{zhao2026selfdistilledreasoner,ye2026opcd,yu2026mopd}.
Visual OPD builds teacher--student asymmetry through evidence-centered crops, recoverable visual cues, or stronger image augmentation for students~\citep{yuan2026visionopd,tian2026vicur,li2026s2vopd}.
Visual contrasts also support supervision gating and token selection~\citep{sun2026vzero,aniri2026opdv}, grouped loss reweighting~\citep{liu2026vaopd}, and target reconstruction~\citep{zhang2026vad}.
Other methods internalize visual manipulation or generated visual thoughts to reduce inference-time computation~\citep{cai2026imagineopd,li2026visualopsd}.
For agents, SGCD~\citep{ding2026sgcd} and GRSD~\citep{zheng2026grsd} use guidance derived from rollout groups to modify credit assignment in reinforcement learning.
\method{} instead extracts reflections from students' visual interactions to reinforce effective behaviors and locate the first failure in \emph{Acquire}, \emph{Read}, or \emph{Ground}.
We condition the teacher on these reflections and use their effects on its predictions to group and reweight tokens for direct distillation.

%% file: contents/03_problem_setup.tex
\section{Preliminaries}
\label{sec:problem-setup}
\paragraph{Agentic visual reasoning.}
Given an image--query pair $x=(I,u)$, a student policy $\pi_\theta$ with parameters $\theta$ interacts with a tool environment $P_{\mathrm{env}}$.
Before step $k$, history $h_k=(x,a_0,o_1,\ldots,a_{k-1},o_k)$ contains the input, previous actions, and returned observations, with $h_0=x$.
An action $a_k$ contains reasoning, executable code, or a final answer.
An observation $o_{k+1}$ contains tool-returned text and images.
The interaction follows
\begin{equation}
    a_k\sim\pi_\theta(\cdot\mid h_k),\qquad
    o_{k+1}\sim P_{\mathrm{env}}(\cdot\mid h_k,a_k),\qquad
    h_{k+1}=(h_k,a_k,o_{k+1}).
    \label{eq:agent-interaction}
\end{equation}
Actions without tool calls yield $o_{k+1}=\emptyset$.
A final answer $\hat y$ ends the interaction, and the complete sequence of actions and observations forms a trajectory $\tau$.

Visual evidence $E_k$ includes $I$ and all tool-returned images in $h_k$.
In Figure~\ref{fig:overview} (bottom), the student crops and zooms in on the blue car's plate.
It reads \mbox{\texttt{S QT 911}} and matches it to option~\texttt{C}.
The correct visual fact $f_x$ needed to answer $u$ provides a reference for checking the student's reading.
The rule $\gamma_x$ maps this fact to the answer.
The evidence chain is
\begin{equation}
    x\xrightarrow{\mathrm{Acquire}}E_k
    \xrightarrow{\mathrm{Read}}f_x
    \xrightarrow[\gamma_x]{\mathrm{Ground}}\hat y.
    \label{eq:evidence-chain}
\end{equation}
Success requires evidence sufficient to establish $f_x$, an accurate reading, and an answer satisfying $\hat y=\gamma_x(f_x)$.
We attribute evidence-chain failures to the first unmet requirement in the dependency order \emph{Acquire}, \emph{Read}, and \emph{Ground}.
These stages can recur throughout an interaction.

\paragraph{On-policy distillation.}
On-policy distillation (OPD)~\citep{agarwal2024onpolicydistillation} trains the student on its own sampled trajectories using token-level teacher distributions.
For each input $x$, the current student samples $M$ trajectories per round, forming $\mathcal G_x=\{\tau_i\}_{i=1}^{M}$ with trajectory index $i$.
Token positions $t$ are distinct from interaction steps $k$.
Token $y_{i,t}$ is the student's output at position $t$ in $\tau_i$.
History $h_{i,t}$ contains the input and preceding student outputs and tool observations.
Let $\mathcal T_i$ denote the supervised student-output positions in trajectory $i$, with $|\mathcal T_i|$ their number.
This set covers reasoning, code, and final answers across interaction rounds.
Prompts and tool returns provide context but are not prediction targets.
The teacher $q_\phi$ has fixed parameters $\phi$.
Given $h_{i,t}$, $p_{i,t}$ and $q^0_{i,t}$ denote the student and base-teacher next-token distributions:
\[
    p_{i,t}=\pi_\theta(\cdot\mid h_{i,t}),\qquad
    q^0_{i,t}=q_\phi(\cdot\mid h_{i,t}).
\]
Superscript $0$ marks the base teacher prediction.
With $\KL$ denoting KL divergence, the reverse-KL objective and its Monte Carlo (MC) estimate from $\mathcal G_x$ are
\begin{equation}
    \mathcal L_{\mathrm{OPD}}(\theta)
    =\mathbb{E}_{\tau_i\sim\pi_\theta(\cdot\mid x)}
    \!\left[\frac{1}{|\mathcal T_i|}
    \sum_{t\in\mathcal T_i}\KL(p_{i,t}\|q^0_{i,t})\right]
    \overset{\mathrm{MC}}{\approx}\frac{1}{M}\sum_{i=1}^{M}\frac{1}{|\mathcal T_i|}
    \sum_{t\in\mathcal T_i}\KL(p_{i,t}\|q^0_{i,t}).
    \label{eq:base-opd}
\end{equation}
The objective averages over supervised positions within each trajectory, then over the $M$ trajectories for the same input.

%% file: contents/04_method.tex
\input{figs/method/figure2.tex}

\section{Method}
\label{sec:method}

\method{} converts same-query student trajectories into reflection-guided
distillation, as shown in Figure~\ref{fig:method-pipeline}.
It first constructs a trajectory-specific visual-evidence reflection,
then measures how the reflection changes a frozen teacher's token
predictions, and finally uses these changes to reweight token-level
distillation losses (Algorithm~\ref{alg:revue-training}).

\subsection{Visual-Evidence Reflection}
\label{sec:reflection}

Our error analysis identifies \emph{Acquire}, \emph{Read}, and \emph{Ground} as the main failure stages (Figure~\ref{fig:overview}), with examples in Appendix~\ref{app:case-study}.
Prior work also shows that attending to the correct visual evidence does not guarantee a correct answer~\citep{liu2026seeing}.
Even with the same evidence, students may need different corrections.
In the pie-chart example in Appendix~\ref{app:read-ground-same-evidence}, both trajectories answer \texttt{No}.
One misreads the slice ranking, while the other reads it correctly but misapplies the lower-median rule.
These failures occur at the \emph{Read} and \emph{Ground} stages.
Targeted distillation guidance therefore requires examining the student's actions, visual observations, and reasoning to identify the specific error in evidence use.

We construct reflections from $\mathcal G_x$, the group of trajectories sampled by the current student for the same input.
The visual evidence and correct reasoning in one attempt can provide a reference for others.
A critic compares these trajectories and their visual observations to extract correct references and diagnose errors in each incorrect trajectory.
For trajectory $i$, the complete visual-evidence reflection is
\begin{equation}
    \mathcal R_i=(\mathcal A_i,\mathcal B_i).
    \label{eq:reflection-components}
\end{equation}
The \textbf{Anchor} $\mathcal A_i$ provides a reference for correct visual evidence use.
It includes the target visual information and supporting images actually observed within $\mathcal G_x$.
It also records the correct visual fact $f_x$ and the rule $\gamma_x$ that maps this fact to the answer.
For an incorrect trajectory, the \textbf{Break Point} $\mathcal B_i=(\sigma_i,\delta_i)$ describes its failure relative to this reference.
$\sigma_i$ denotes the first failed stage in the dependency order, and $\delta_i$ describes the specific discrepancy.
These images and text serve as training-time teacher context for token-level supervision on the student's sampled trajectories.
The teacher can then reinforce effective evidence use and guide corrections to observed errors.
Appendix~\ref{app:reflection-case} shows a concrete visual-evidence reflection, and Appendices~\ref{app:critic-feedback}--\ref{app:critic-prompts} detail its construction and critic prompts.

\input{figs/method/figure3.tex}

\subsection{From Reflection to Token Impact}
\label{sec:token-impact}
Visual-evidence reflection provides correct references and diagnoses failures in the evidence chain.
To locate where this guidance changes supervision, we compare teacher scores along the student's sampled trajectory.
On the same student history $h_{i,t}$, the frozen teacher predicts $q^0_{i,t}$ without reflection (Section~\ref{sec:problem-setup}) and $q^{\mathcal R}_{i,t}=q_\phi(\cdot\mid h_{i,t},\mathcal R_i)$ with it.
Only the training-time teacher receives reflection; the student retains its original interaction history.
Using teacher probabilities normalized over the full vocabulary, we define the log-probability shift for candidate token $v$:
\begin{equation}
    d_{i,t}(v)=\log q^{\mathcal R}_{i,t}(v)-\log q^0_{i,t}(v).
    \label{eq:teacher-shift}
\end{equation}
To connect these shifts to the reverse-KL objective in \Eqref{eq:base-opd}, we compare both teachers on a shared vocabulary support.
The training loss uses the adopted teacher's top-$K$ candidates; for the reflection-conditioned branch, we use these candidates as $\mathrm{Supp}$ for both teachers.
Suppressing $(i,t)$, write $p=p_{i,t}$ and $p^{\mathrm{Supp}}(v)=p(v)/\sum_{u\in\mathrm{Supp}}p(u)$ for $v\in\mathrm{Supp}$.
Define $q^{\mathrm{Supp}}$ analogously and let $\ell^{\mathrm{Supp}}(p,q)=\KL(p^{\mathrm{Supp}}\|q^{\mathrm{Supp}})$.
\begin{lemma}[Reflection and the distillation gradient]
\label{lem:reflection-gradient}
For fixed history, teacher predictions, and nonempty shared support, with positive student and teacher probabilities on this support, we have
\begin{equation}
    \nabla_\theta\!\left[\ell^{\mathrm{Supp}}(p,q^{\mathcal R})-\ell^{\mathrm{Supp}}(p,q^0)\right]
    =-\mathbb{E}_{v\sim p^{\mathrm{Supp}}}\!\left[d(v)\nabla_\theta\log p^{\mathrm{Supp}}(v)\right].
    \label{eq:reflection-gradient}
\end{equation}
\end{lemma}
\Eqref{eq:reflection-gradient} expresses the gradient change as a sum over candidate tokens, with each term weighted by its teacher-score shift $d(v)$ (proof in Appendix~\ref{app:reflection-proof}).
We focus on the student's sampled token $y_{i,t}$ to locate \emph{where reflection changes teacher support for its actual choices}.
We define the magnitude of this shift as \textbf{\emph{token impact}}:
\begin{equation}
    s_{i,t}=|d_{i,t}(y_{i,t})|
    =\left|\log q^{\mathcal R}_{i,t}(y_{i,t})-\log q^0_{i,t}(y_{i,t})\right|.
    \label{eq:token-impact}
\end{equation}
Taking the absolute value captures both increased and decreased teacher support, covering reinforcement and correction.
This score ranks positions by changes in teacher scores, not by full gradient magnitude.
Each position's loss still supervises the distribution over its support.

Token impact distinguishes positions within a trajectory by how much their teacher scores change with reflection.
In Figure~\ref{fig:token-impact-case}, reflection reduces support for the mistaken judgment of \texttt{one} island.
The token \texttt{one} has higher impact than the final \texttt{Lucia}, whose teacher score changes little under the existing erroneous prefix.
Section~\ref{sec:grouped-distillation} groups positions by these scores and assigns greater distillation weight to high-impact positions.

\subsection{Reflection-Aware Grouped Distillation}
\label{sec:grouped-distillation}

\begingroup
\input{figs/method/figure4_sparsity.tex}

Token impact measures teacher-score changes induced by visual-evidence reflection (Section~\ref{sec:token-impact}).
Figure~\ref{fig:token-impact-sparsity} pools early-training scores: approximately 20\% of scored positions account for 97.6\% of total impact (Appendix~\ref{app:token-impact-sparsity}).
Uniform token averaging allocates total loss weight in proportion to group size, leaving sparse high-impact positions a small share.
To emphasize their supervision, we group positions by impact and introduce weights $w_{i,t}$ into the reverse-KL objective in \Eqref{eq:base-opd}:
\par
\endgroup
\begin{equation}
\mathcal L_{\mathrm{ReVuE}}(\theta)
=
\frac{1}{|\mathcal G|}
\sum_{\tau_i\in\mathcal G}
\frac{1}{|\mathcal T_i|}
\sum_{t\in\mathcal T_i}
\textcolor{red}{w_{i,t}}\,
\KL\!\left(
p_{i,t}^{\mathrm{Supp}_{i,t}}
\,\middle\|\,
(q^*_{i,t})^{\mathrm{Supp}_{i,t}}
\right).
\label{eq:grouped-objective}
\end{equation}
$\mathcal G$ contains retained student trajectories in the current batch, with nonempty supervised sets $\mathcal T_i$.
The teacher target $q^*_{i,t}$ is $q^{\mathcal R}_{i,t}$ when reflection-conditioned scoring is valid and token-aligned, and $q^0_{i,t}$ otherwise.
Both distributions are renormalized over $\mathrm{Supp}_{i,t}=\operatorname{TopK}(q^*_{i,t})$, as indicated by the superscripts.

Within each trajectory, we rank supervised positions by $s_{i,t}$, assigning the top $\lceil\alpha|\mathcal T_i|\rceil$ to $\mathcal T_{i,\mathrm{High}}$ and the rest to $\mathcal T_{i,\mathrm{Low}}=\mathcal T_i\setminus\mathcal T_{i,\mathrm{High}}$, where $\alpha\in(0,1)$.
For nonempty groups, we allocate total loss weights $\lambda\in(0,1)$ and $1-\lambda$ to \texttt{High} and \texttt{Low} and distribute each uniformly within its group.
Each position's coefficient in \Eqref{eq:grouped-objective} is $w_{i,t}/|\mathcal T_i|$, giving the weights
\begin{equation}
    w_{i,t}=
    \begin{cases}
        \lambda |\mathcal T_i|/|\mathcal T_{i,\mathrm{High}}|,
        & t\in\mathcal T_{i,\mathrm{High}},\\
        (1-\lambda)|\mathcal T_i|/|\mathcal T_{i,\mathrm{Low}}|,
        & t\in\mathcal T_{i,\mathrm{Low}}.
    \end{cases}
    \label{eq:token-weights}
\end{equation}
$\alpha$ controls group size, while $\lambda$ controls total group weight.
We set $\lambda=0.5$, assigning half to each group.
With teacher targets, supports, and groups fixed, $w_{i,t}$ scales each position's gradient relative to uniform token averaging.
This multiplier exceeds one for \texttt{High} when the group contains fewer than half the supervised positions.
Appendix~\ref{app:group-weight-scaling} and Figure~\ref{fig:group-weight-scaling} give the derivation and weight curves.
If impacts are unavailable or either group is empty, we restore uniform averaging with $w_{i,t}=1$ and retain the adopted teacher target (Appendix~\ref{app:training-procedure}).

%% file: figs/method/figure2.tex
\begin{figure}[!t]
  \centering
  \includegraphics[width=\textwidth]{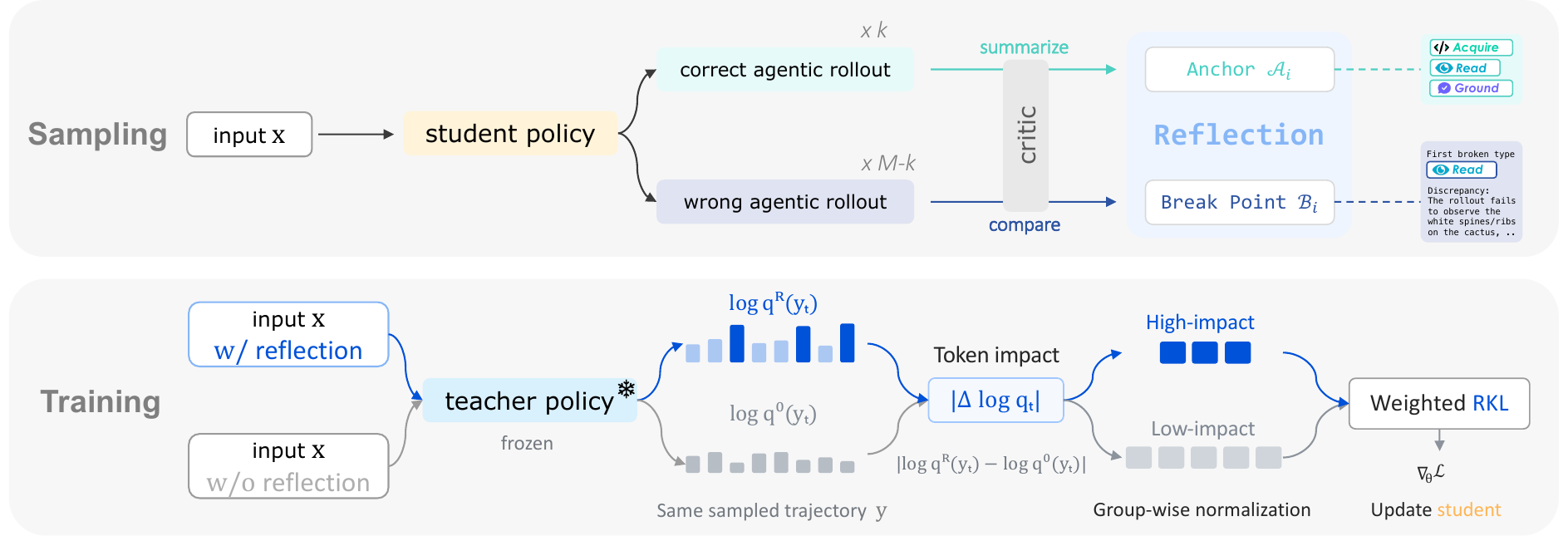}
  \caption{
  \textbf{Sampling and training pipeline of \method{}.}
  }
  \label{fig:method-pipeline}
\end{figure}

%% file: figs/method/figure3.tex
\begin{figure}[!t]
  \centering
  \includegraphics[width=\textwidth]{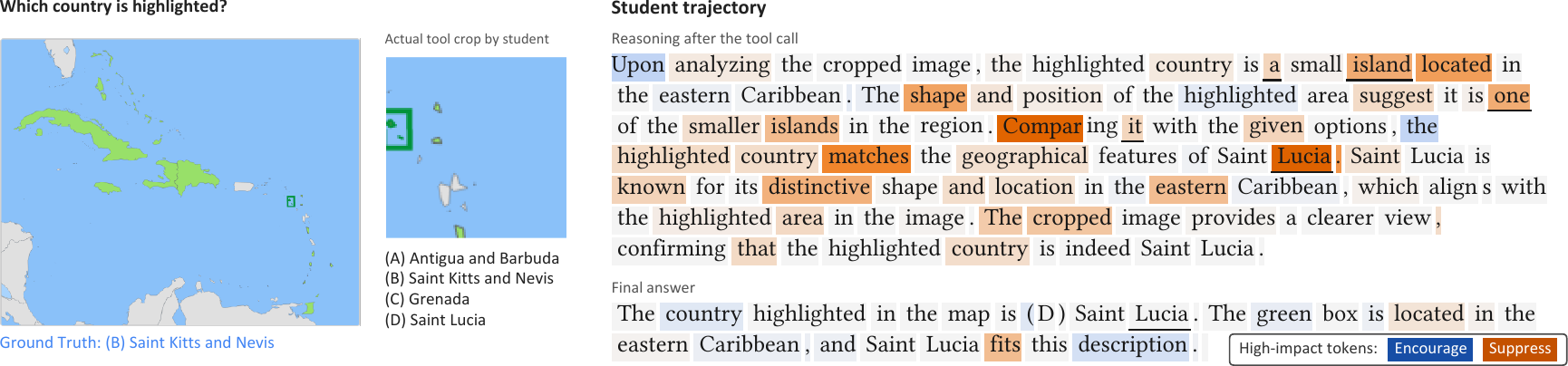}
  \caption{
  \textbf{Token impact on a trajectory with a \textnormal{\emph{reading}} error.}
  The student misreads two islands as one, answering \texttt{Saint Lucia} instead of \texttt{Saint Kitts and Nevis}.
  Reflection identifies two islands and lowers support for the \underline{underlined} \texttt{island}, \texttt{one}, and first \texttt{Lucia}.
  \textcolor{tokenImpactBlue}{\textbf{Blue}}/\textcolor{tokenImpactOrange}{\textbf{orange}} denote increased/decreased teacher support; darker shading indicates larger absolute log-probability shifts.
  Both evaluations score the same student trajectory; the final \texttt{Lucia} changes little under the fixed erroneous prefix.
  }
  \label{fig:token-impact-case}
\end{figure}

%% file: figs/method/figure4_sparsity.tex
\begin{wrapfigure}{r}{0.40\textwidth}
  \centering
  \includegraphics[width=\linewidth]{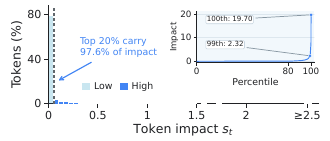}
  \par\nointerlineskip
  \caption{
  \textbf{Token-impact sparsity}
  }
  \label{fig:token-impact-sparsity}
\end{wrapfigure}

%% file: contents/05_experiments.tex
\section{Experiments}
\label{sec:experiments}

\subsection{Experimental Setup}
\label{sec:experimental-setup}

\textbf{For models,} we use Thyme-SFT~\citep{zhang2026thyme} cold-start checkpoints of Qwen2.5-VL-7B~\citep{bai2025qwen25vl} and InternVL3.5-4B-Instruct~\citep{wang2025internvl35} as students.
Teachers are our reproduced Thyme-RL experts; Qwen3.5-397B-A17B~\citep{qwen2026qwen35} is the critic.
Off-the-shelf comparisons include GPT-4o~\citep{openai2024gpt4o}, Gemini-3.1-Flash-Lite, Qwen2.5-VL-32B, Qwen3-VL-30B-A3B-Thinking~\citep{bai2025qwen3vl}, and InternVL3.5-38B-Instruct.
\textbf{For baselines,} we compare with Vision-OPD~\citep{yuan2026visionopd} for visual privilege and V-Zero~\citep{sun2026vzero} and VAD~\citep{zhang2026vad} for visual contrast.
We also include Vanilla OPD, RFT~\citep{yuan2023scaling}, and GT-Privileged.
All OPD baselines use the same RL expert teacher; details are shown in Appendix~\ref{app:baseline-details}.
\textbf{For benchmarks,} we evaluate on 11 benchmarks covering perception, math, and general multimodal tasks (Appendix~\ref{app:benchmarks}).
Appendix~\ref{app:training-evaluation-settings} details the training data, model initialization, and training and evaluation settings.

\subsection{Main Results}
\label{sec:main-results}
\begin{TakeawayBox}
\input{contents/main_results_takeaways.tex}
\end{TakeawayBox}

\input{figs/perception_results_20260925/main_figure.tex}
\textbf{\method{} improves perception across model families, with gains in math and general tasks.}
In all three weighted category averages, both \method{} students outperform all evaluated OPD baselines and RFT, which trains only on teacher trajectories with correct answers (Table~\ref{tab:main-results}).
Perception gains over Vanilla OPD are 2.40\% for Qwen2.5-VL-7B and 1.70\% for InternVL3.5-4B-Instruct.
Qwen's gains on HRBench 8K (+3.50\%) and TreeBench (+3.10\%) cover high-resolution understanding and object relations.
Both models also improve in math and general tasks, with average gains of 2.83\% and 1.65\% for InternVL, respectively.
\method{} also achieves higher category averages than baselines using privileged visual information or visual contrasts, with matched student initializations and teacher sources.
The answer-privilege comparison provides further evidence: \method{} exceeds GT-Privileged on all 11 benchmarks for both students.
These comparisons support guiding how students acquire, read, and use evidence; giving the teacher the correct answer alone does not yield the same gains.
After reflection-assisted distillation, both students also surpass their RL Expert teachers in all three category averages.
Both students also exceed GPT-4o in all three averages and the larger off-the-shelf models in their families in perception.
\input{tables/main_results.tex}

\textbf{\method{} improves perception accuracy with shorter responses and less frequent tool use.}
Across all five perception benchmarks, the Qwen2.5-VL-7B student is more accurate than Vanilla OPD while producing shorter responses and using tools less often (Figure~\ref{fig:perception-efficiency}; Appendix~\ref{app:perception-efficiency}).
On HRBench 8K, \method{} reduces mean response length from the RL Expert's 458 tokens to 382, while raising accuracy from 71.75\% to 74.00\% (+2.25\%).
On HRBench 8K and V*~Bench, its answer accuracy exceeds the other plotted distillation baselines on samples both with and without tool use.
Tool-use frequency alone does not explain the performance differences.
On V*~Bench, VAD uses tools slightly less often (73.3\% vs.\ 74.3\%), yet \method{} achieves 5.24\% higher overall accuracy.
The extent of these behavioral changes varies by task.
On TreeBench, \method{} reduces response length by 46.09\% and tool-use rate from 64.7\% to 15.8\%, while improving accuracy by 3.10\% over Vanilla OPD.
On VisualProbe, it still uses tools on 83.7\% of samples and also achieves higher accuracy.
These results support more effective visual evidence use: higher accuracy with less generation and less frequent tool use, while tool reliance still varies by task.

%% file: contents/main_results_takeaways.tex
\begin{enumerate}
\setlength{\itemsep}{0.45em}
\setlength{\parsep}{0pt}
\item \textbf{Both model families improve across all three task categories.}
Both \method{} students exceed the evaluated OPD baselines and RFT in all three weighted category averages.
Perception gains over Vanilla OPD are 2.40\% for Qwen2.5-VL-7B and 1.70\% for InternVL3.5-4B-Instruct.

\item \textbf{Higher perception accuracy accompanies shorter responses and less frequent tool use.}
Qwen2.5-VL-7B improves all three measures over Vanilla OPD on all five perception benchmarks, with the extent of change varying by task.
\end{enumerate}

%% file: figs/perception_results_20260925/main_figure.tex
\begin{figure}[!htbp]
  \centering
  \includegraphics[width=0.9\linewidth]{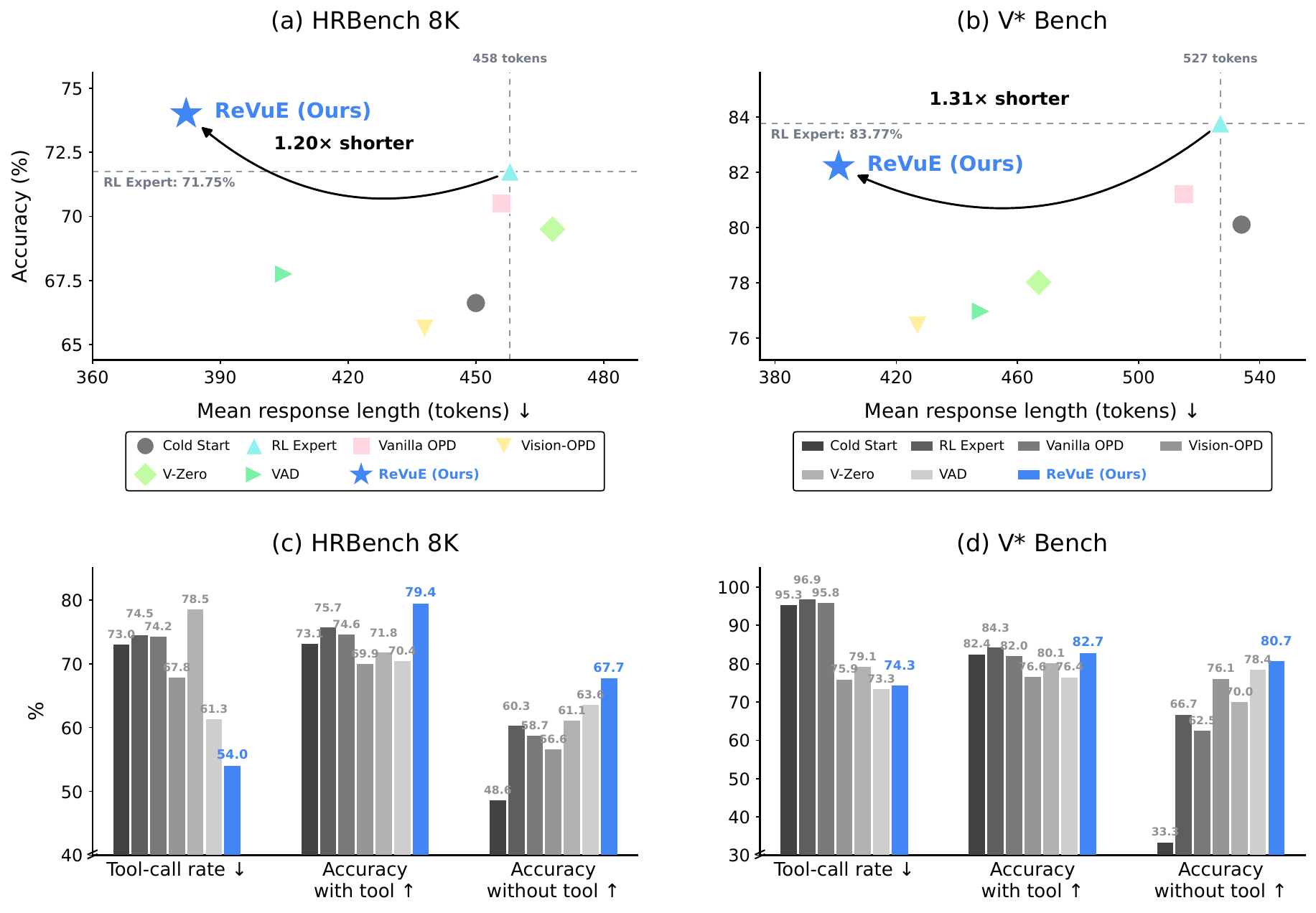}
  \caption{
  \textbf{Response length, accuracy, and tool use} for Qwen2.5-VL-7B.
  }
  \label{fig:perception-efficiency}
\end{figure}

%% file: tables/main_results.tex
\begin{table}[t]
  \centering
  \setlength{\belowcaptionskip}{6pt}
  \caption{
  Main results.
  \textbf{Bold} marks the best OPD result per column within each model family, including ties.
  Wtd. Avg. uses benchmark sample counts as weights.
  Baseline details and model labels are provided in Appendix~\ref{app:baseline-details}--\ref{app:model-labels}.
  Evaluation benchmark details are provided in Appendix~\ref{app:benchmarks}.
  }
  \label{tab:main-results}
  \begingroup
  \fontsize{8.5}{10}\selectfont
  \renewcommand{\arraystretch}{1.15}
  \setlength{\tabcolsep}{1.3pt}
  \setlength{\arrayrulewidth}{0.25pt}
  \newlength{\mainmetricwidth}
  \setlength{\mainmetricwidth}{\dimexpr(\linewidth-55pt-29\tabcolsep-6\arrayrulewidth)/14\relax}
    \begin{tabular}{@{}p{55pt}
      *{5}{>{\centering\arraybackslash}p{\mainmetricwidth}}|>{\centering\arraybackslash}p{\mainmetricwidth}|
      *{3}{>{\centering\arraybackslash}p{\mainmetricwidth}}|>{\centering\arraybackslash}p{\mainmetricwidth}|
      *{3}{>{\centering\arraybackslash}p{\mainmetricwidth}}|>{\centering\arraybackslash}p{\mainmetricwidth}|@{}}
      \toprule
      \multirow{2}{55pt}[-4pt]{\centering\bfseries Model /\\Method}
      & \multicolumn{6}{c}{\textbf{Perception}}
      & \multicolumn{4}{c}{\textbf{Math}}
      & \multicolumn{4}{c}{\textbf{General}} \\
      \cmidrule(lr){2-7}\cmidrule(lr){8-11}\cmidrule(lr){12-15}
      & {\fontsize{7.5}{8.5}\selectfont\shortstack{HR-4K}}
      & {\fontsize{7.5}{8.5}\selectfont\shortstack{HR-8K}}
      & {\fontsize{7.5}{8.5}\selectfont\shortstack{V\textsuperscript{*}}}
      & {\fontsize{7.5}{8.5}\selectfont\shortstack{Tree\\Bench}}
      & {\fontsize{7.5}{8.5}\selectfont\shortstack{Visual\\Probe}}
      & {\fontsize{7.5}{8.5}\selectfont\shortstack{Wtd.\\Avg.}}
      & {\fontsize{7.5}{8.5}\selectfont\shortstack{Math\\Vista}}
      & {\fontsize{7.5}{8.5}\selectfont\shortstack{Math\\Verse}}
      & {\fontsize{7.5}{8.5}\selectfont\shortstack{Visu\\Logic}}
      & {\fontsize{7.5}{8.5}\selectfont\shortstack{Wtd.\\Avg.}}
      & {\fontsize{7.5}{8.5}\selectfont\shortstack{Hallu\\Bench}}
      & {\fontsize{7.5}{8.5}\selectfont\shortstack{CQA\\Pro}}
      & {\fontsize{7.5}{8.5}\selectfont\shortstack{Info\\VQA}}
      & {\fontsize{7.5}{8.5}\selectfont\shortstack{Wtd.\\Avg.}} \\
      \midrule
      \rowcolor{tableModelGray}
      \multicolumn{15}{c}{\mainTableModelHeader{Off-the-Shelf Models}} \\
      GPT-4o & 61.00 & 54.00 & 61.78 & 49.88 & 23.88 & 50.28 & 58.83 & 39.21 & 25.20 & 41.22 & 51.37 & 28.67 & 71.17 & 53.28 \\
      Gemini3.1FL & 46.00 & 43.00 & 64.92 & 51.85 & 27.96 & 43.89 & 80.80 & 77.92 & 32.40 & 62.63 & 59.92 & 37.03 & 83.50 & 63.57 \\
      \mbox{Qwen2.5-32B} & 75.13 & 69.25 & 78.01 & 48.40 & 45.05 & 63.89 & 77.00 & 53.05 & 25.90 & 51.90 & 50.74 & 30.02 & 83.10 & 59.29 \\
      \mbox{Qwen3-30B-T} & 77.13 & 71.38 & 80.10 & 45.43 & 36.89 & 63.26 & 80.20 & 66.12 & 25.80 & 56.71 & 61.82 & 35.46 & 85.68 & 64.45 \\
      \mbox{InternVL-38B} & 71.50 & 62.13 & 65.97 & 41.98 & 20.97 & 54.34 & 70.90 & 48.48 & 27.20 & 48.89 & 50.74 & 26.98 & 77.70 & 55.71 \\
      \midrule
      \rowcolor{tableModelGray}
      \multicolumn{15}{c}{\mainTableModelHeader{Qwen2.5-VL-7B}} \\
      \textcolor{tableColdGray}{Base Model} & \textcolor{tableColdGray}{69.00} & \textcolor{tableColdGray}{63.50} & \textcolor{tableColdGray}{75.39} & \textcolor{tableColdGray}{37.04} & \textcolor{tableColdGray}{41.17} & \textcolor{tableColdGray}{57.77} & \textcolor{tableColdGray}{69.10} & \textcolor{tableColdGray}{44.04} & \textcolor{tableColdGray}{25.40} & \textcolor{tableColdGray}{46.34} & \textcolor{tableColdGray}{40.91} & \textcolor{tableColdGray}{20.79} & \textcolor{tableColdGray}{75.10} & \textcolor{tableColdGray}{50.53} \\
      \textcolor{tableColdGray}{Cold-start} & \textcolor{tableColdGray}{74.38} & \textcolor{tableColdGray}{66.62} & \textcolor{tableColdGray}{80.10} & \textcolor{tableColdGray}{37.28} & \textcolor{tableColdGray}{41.56} & \textcolor{tableColdGray}{60.72} & \textcolor{tableColdGray}{68.00} & \textcolor{tableColdGray}{44.29} & \textcolor{tableColdGray}{26.40} & \textcolor{tableColdGray}{46.38} & \textcolor{tableColdGray}{41.73} & \textcolor{tableColdGray}{21.12} & \textcolor{tableColdGray}{79.05} & \textcolor{tableColdGray}{52.68} \\
      \textcolor{tableExpertGray}{RL Expert} & \textcolor{tableExpertGray}{75.50} & \textcolor{tableExpertGray}{71.75} & \textcolor{tableExpertGray}{83.77} & \textcolor{tableExpertGray}{40.25} & \textcolor{tableExpertGray}{44.85} & \textcolor{tableExpertGray}{63.89} & \textcolor{tableExpertGray}{70.60} & \textcolor{tableExpertGray}{45.43} & \textcolor{tableExpertGray}{26.20} & \textcolor{tableExpertGray}{47.56} & \textcolor{tableExpertGray}{43.83} & \textcolor{tableExpertGray}{21.69} & \textcolor{tableExpertGray}{79.74} & \textcolor{tableExpertGray}{53.60} \\
      \cmidrule{1-15}
      \textcolor{tableExpertGray}{RFT} & \textcolor{tableExpertGray}{75.38} & \textcolor{tableExpertGray}{72.00} & \textcolor{tableExpertGray}{81.20} & \textcolor{tableExpertGray}{40.00} & \textcolor{tableExpertGray}{41.75} & \textcolor{tableExpertGray}{63.12} & \textcolor{tableExpertGray}{69.50} & \textcolor{tableExpertGray}{45.81} & \textcolor{tableExpertGray}{24.60} & \textcolor{tableExpertGray}{46.70} & \textcolor{tableExpertGray}{42.33} & \textcolor{tableExpertGray}{21.18} & \textcolor{tableExpertGray}{79.57} & \textcolor{tableExpertGray}{53.07} \\
      Vanilla OPD & 75.40 & 70.50 & 81.20 & 38.02 & 42.91 & 62.61 & 70.60 & 46.07 & 26.00 & 47.67 & 42.88 & 21.69 & 79.44 & 53.28 \\
      \mbox{GT-Privileged} & 73.62 & 70.50 & 79.58 & 39.75 & 43.10 & 62.26 & 71.20 & 45.69 & 25.20 & 47.49 & 43.45 & 20.84 & 79.70 & 53.23 \\
      Vision-OPD & 73.00 & 65.62 & 76.44 & 39.01 & 41.36 & 59.98 & 67.70 & 42.39 & 25.70 & 45.48 & 39.89 & 20.79 & 78.75 & 52.08 \\
      V-Zero & 75.62 & 69.50 & 78.01 & 37.28 & 43.10 & 62.08 & 69.80 & 43.91 & \textbf{26.60} & 46.99 & 42.16 & 21.08 & 79.12 & 52.79 \\
      VAD & 74.75 & 67.75 & 76.96 & \textbf{41.98} & 43.89 & 62.08 & 69.00 & 45.56 & 22.30 & 45.62 & 44.24 & 21.73 & 79.65 & 53.65 \\
      \rowcolor{tableOursBlue}
      \textbf{\ours{}} & \textbf{77.10} & \textbf{74.00} & \textbf{82.20} & 41.12 & \textbf{44.66} & \textbf{65.01} & \textbf{71.60} & \textbf{47.08} & 26.50 & \textbf{48.49} & \textbf{45.12} & \textbf{21.81} & \textbf{80.27} & \textbf{54.14} \\
      \midrule
      \rowcolor{tableModelGray}
      \multicolumn{15}{c}{\mainTableModelHeader{InternVL3.5-4B-Instruct}} \\
      \textcolor{tableColdGray}{Base Model} & \textcolor{tableColdGray}{62.00} & \textcolor{tableColdGray}{55.00} & \textcolor{tableColdGray}{68.59} & \textcolor{tableColdGray}{40.49} & \textcolor{tableColdGray}{20.58} & \textcolor{tableColdGray}{49.32} & \textcolor{tableColdGray}{68.50} & \textcolor{tableColdGray}{28.93} & \textcolor{tableColdGray}{25.80} & \textcolor{tableColdGray}{42.00} & \textcolor{tableColdGray}{40.04} & \textcolor{tableColdGray}{26.01} & \textcolor{tableColdGray}{66.04} & \textcolor{tableColdGray}{47.78} \\
      \textcolor{tableColdGray}{Cold-start} & \textcolor{tableColdGray}{69.50} & \textcolor{tableColdGray}{63.25} & \textcolor{tableColdGray}{66.49} & \textcolor{tableColdGray}{40.99} & \textcolor{tableColdGray}{29.90} & \textcolor{tableColdGray}{55.66} & \textcolor{tableColdGray}{69.30} & \textcolor{tableColdGray}{47.21} & \textcolor{tableColdGray}{25.80} & \textcolor{tableColdGray}{47.45} & \textcolor{tableColdGray}{48.50} & \textcolor{tableColdGray}{26.24} & \textcolor{tableColdGray}{72.98} & \textcolor{tableColdGray}{52.79} \\
      \textcolor{tableExpertGray}{RL Expert} & \textcolor{tableExpertGray}{72.62} & \textcolor{tableExpertGray}{64.62} & \textcolor{tableExpertGray}{71.73} & \textcolor{tableExpertGray}{40.74} & \textcolor{tableExpertGray}{34.56} & \textcolor{tableExpertGray}{58.20} & \textcolor{tableExpertGray}{69.30} & \textcolor{tableExpertGray}{46.32} & \textcolor{tableExpertGray}{26.90} & \textcolor{tableExpertGray}{47.60} & \textcolor{tableExpertGray}{46.92} & \textcolor{tableExpertGray}{30.59} & \textcolor{tableExpertGray}{73.93} & \textcolor{tableExpertGray}{54.38} \\
      \cmidrule{1-15}
      \textcolor{tableExpertGray}{RFT} & \textcolor{tableExpertGray}{71.25} & \textcolor{tableExpertGray}{66.00} & \textcolor{tableExpertGray}{71.73} & \textcolor{tableExpertGray}{40.25} & \textcolor{tableExpertGray}{35.00} & \textcolor{tableExpertGray}{58.22} & \textcolor{tableExpertGray}{70.00} & \textcolor{tableExpertGray}{45.94} & \textcolor{tableExpertGray}{24.30} & \textcolor{tableExpertGray}{46.81} & \textcolor{tableExpertGray}{47.78} & \textcolor{tableExpertGray}{29.39} & \textcolor{tableExpertGray}{74.17} & \textcolor{tableExpertGray}{54.26} \\
      Vanilla OPD & 71.13 & 65.87 & \textbf{73.30} & 40.25 & 33.79 & 58.02 & 68.40 & 43.40 & 26.60 & 46.34 & 46.85 & 30.46 & 72.16 & 53.48 \\
      \mbox{GT-Privileged} & 71.25 & 64.88 & 70.16 & 39.75 & 33.20 & 57.36 & 71.30 & 39.34 & 26.20 & 46.09 & 49.29 & 29.90 & 72.48 & 53.91 \\
      Vision-OPD & 66.25 & 60.00 & 65.97 & 37.78 & 28.93 & 53.04 & 66.60 & 44.54 & 26.60 & 46.02 & 44.57 & 30.46 & \textbf{74.47} & 54.14 \\
      V-Zero & 68.13 & 61.75 & 67.02 & 40.25 & 28.35 & 54.45 & 68.80 & 46.95 & 26.80 & 47.56 & 49.03 & 28.65 & 73.61 & 53.99 \\
      VAD & 67.50 & 61.00 & 67.02 & 39.01 & 27.96 & 53.78 & 68.30 & 46.07 & 27.70 & 47.45 & 48.05 & 31.44 & 73.36 & 54.61 \\
      \rowcolor{tableOursBlue}
      \textbf{\ours{}} & \textbf{73.25} & \textbf{67.37} & \textbf{73.30} & \textbf{41.48} & \textbf{36.12} & \textbf{59.72} & \textbf{71.60} & \textbf{47.46} & \textbf{28.10} & \textbf{49.17} & \textbf{49.38} & \textbf{32.27} & 73.35 & \textbf{55.13} \\
      \bottomrule
    \end{tabular}
  \endgroup
\end{table}

%% file: contents/06_analysis_and_discussion.tex
\section{Analysis and Discussion}
\label{sec:analysis}
\begin{TakeawayBox}
\input{contents/analysis_takeaways.tex}
\end{TakeawayBox}

\ReportNeedspace{10\baselineskip}
\begingroup
\setlength{\intextsep}{3pt}
\setlength{\columnsep}{10pt}
\input{tables/core_ablation.tex}
\paragraph{Reflection brings gains across tasks; impact reweighting further improves perception and math.}
In the cumulative ablation on Qwen2.5-VL-7B, adding visual-evidence reflection improves all three category averages and nine of 11 benchmark scores over Vanilla OPD (Table~\ref{tab:core-ablation}).
Adding impact reweighting then raises the perception and math averages by another 1.31\% and 0.71\%, respectively.
It also improves tasks that do not benefit from reflection alone.
On TreeBench, for example, accuracy falls from 38.02\% to 37.78\% with reflection alone, then rises to 41.12\% with reweighting.
The general-task average drops by 0.28\% relative to reflection alone, but \method{} still outperforms Vanilla OPD on all 11 benchmarks (Appendix~\ref{app:component-ablation}).
\par
\endgroup

\ReportNeedspace{10\baselineskip}
\begingroup
\setlength{\intextsep}{2pt}
\setlength{\columnsep}{8pt}
\input{tables/impact_intervention.tex}
\paragraph{Impact ranking identifies supervision worth prioritizing.}
We compare impact-based and random selection on Qwen2.5-VL-7B with fixed group fractions and weights (Table~\ref{tab:impact-intervention}).
Compared with random selection, \method{} gains 1.37\% in perception and 1.07\% in math, but only 0.04\% on general tasks.
Masking yields the same ordering in all three category averages: \texttt{Mask low} $>$ \texttt{Mask random} $>$ \texttt{Mask top}.
\texttt{Mask low} also exceeds \texttt{Mask top} on all 11 benchmarks (Appendix~\ref{app:impact-intervention}).
The cases show reflection lowering teacher support for erroneous choices and raising support for correct judgments within failed trajectories.
In the age case, reflection favors the correct relation \texttt{above} and discourages the erroneous comparison \texttt{exceed} (Appendix~\ref{app:token-impact-cases}).
Scores can change at target selection and intermediate visual judgments while final-answer token scores remain nearly unchanged (Figure~\ref{fig:token-impact-case}; Appendix~\ref{app:token-impact-cases}).
The top 20\% of pooled early-training positions carry 97.6\% of total impact, supporting greater weight on these positions (Figure~\ref{fig:token-impact-sparsity}).
\texttt{Mask low} gains 0.17\% in perception over the full method but slightly reduces math and general averages, showing task-dependent effects.
\par
\endgroup

\ReportNeedspace{10\baselineskip}
\begingroup
\setlength{\intextsep}{2pt}
\setlength{\columnsep}{8pt}
\input{tables/alpha_sensitivity.tex}
\paragraph{Prioritizing 20\% of positions gives the highest perception and math averages.}
This fraction also leads on all three math benchmarks (Table~\ref{tab:alpha-sensitivity}; Appendix~\ref{app:alpha-sensitivity}).
With $\lambda=0.5$, reducing $\alpha$ shrinks the high-weight group and increases each position's weight within it (Appendix~\ref{app:group-weight-scaling}).
Neither 5\% nor 10\% exceeds the perception and math averages at 20\%, supporting moderate concentration in this setting.
The general-task average peaks at 100\%, 0.28\% above the 20\% setting, indicating different preferences across tasks.
\par
\endgroup

\ReportNeedspace{10\baselineskip}
\begingroup
\setlength{\intextsep}{3pt}
\setlength{\columnsep}{9pt}
\input{tables/critic_consistency.tex}
\paragraph{Critic diagnoses agree closely with human judgments, including on incorrect trajectories.}
One annotator independently labels 96 rollouts balanced across \texttt{all-correct}, \texttt{mixed}, and \texttt{all-wrong} groups before viewing critic labels.
Final stage labels match 93 of 96 human labels (96.88\%, Cohen's $\kappa=0.957$).
Agreement remains 95.31\% on the 64 incorrect rollouts.
On 768 rollouts, we compare archived final training labels with fresh labels from the same critic and two external judges.
Valid pairs agree at 99.2\% for the same critic and 92.8--93.2\% across models (Table~\ref{tab:critic-consistency}).
These results support human agreement and repeatability of the stage diagnoses used for reflection.
Appendix~\ref{app:critic-consistency} details sampling and label disagreements.
\par
\endgroup

%% file: contents/analysis_takeaways.tex
\begin{enumerate}
\setlength{\itemsep}{0.45em}
\setlength{\parsep}{0pt}
\item \textbf{Reflection improves all three category averages; reweighting adds gains in perception and math.}
On Qwen2.5-VL-7B, reflection raises all three averages, and adding impact reweighting raises perception and math by another 1.31\% and 0.71\%, respectively.
The general-task average drops by 0.28\% relative to reflection alone.

\item \textbf{Impact ranking identifies supervision worth prioritizing.}
On Qwen2.5-VL-7B, impact-based selection outperforms random selection in perception and math, and masking high-impact positions hurts more than masking equally many low-impact positions.
The top 20\% of pooled early-training positions carry 97.6\% of total impact.

\item \textbf{Prioritizing 20\% of positions gives the best tested perception and math averages.}
General tasks on Qwen2.5-VL-7B favor the 100\% setting, whose average exceeds the 20\% setting by 0.28\%.

\item \textbf{Critic stage diagnoses agree closely with human annotation and other judges.}
Critic labels match one annotator on 93 of 96 rollouts (96.88\%), with 95.31\% agreement on the 64 incorrect rollouts.
Valid pairs agree at 99.2\% for a critic rerun and 92.8--93.2\% across models.
\end{enumerate}

%% file: tables/core_ablation.tex
\begin{wraptable}[6]{r}{0.56\textwidth}
  \centering
  \setlength{\abovecaptionskip}{0pt}
  \setlength{\belowcaptionskip}{6pt}
  \caption{
  \textbf{Component ablation on Qwen2.5-VL-7B.}
  }
  \label{tab:core-ablation}
  \begingroup
  \footnotesize
  \renewcommand{\arraystretch}{1.12}
  \setlength{\tabcolsep}{3pt}
  \setlength{\arrayrulewidth}{0.4pt}
  \begin{tabular}{lccc}
    \textbf{Method} & \textbf{Perception} $\uparrow$ & \textbf{Math} $\uparrow$ & \textbf{General} $\uparrow$ \\
    \hline
    \singleRuleTableGap
    \textcolor{tableColdGray}{Vanilla OPD} & \textcolor{tableColdGray}{62.61} & \textcolor{tableColdGray}{47.67} & \textcolor{tableColdGray}{53.28} \\
    \hspace*{0.4em}\textbf{+reflection} & 63.70 & 47.78 & \textbf{54.42} \\
    \rowcolor{tableOursBlue}
    \hspace*{0.8em}\textbf{+impact reweight} & \textbf{65.01} & \textbf{48.49} & 54.14 \\
  \end{tabular}
  \endgroup
\end{wraptable}

%% file: tables/impact_intervention.tex
\begin{wraptable}{r}{0.445\textwidth}
  \centering
  \setlength{\abovecaptionskip}{0pt}
  \setlength{\belowcaptionskip}{2pt}
  \caption{
  \textbf{Token-impact interventions.}
  }
  \label{tab:impact-intervention}
  \begingroup
  \fontsize{8}{9.2}\selectfont
  \renewcommand{\arraystretch}{1.02}
  \setlength{\tabcolsep}{1.8pt}
  \setlength{\arrayrulewidth}{0.4pt}
  \begin{tabular}{@{}lccc@{}}
    \textbf{Method} & \textbf{Perception} $\uparrow$ & \textbf{Math} $\uparrow$ & \textbf{General} $\uparrow$ \\
    \hline
    \singleRuleTableGap
    \textcolor{tableColdGray}{Vanilla OPD} & \textcolor{tableColdGray}{62.61} & \textcolor{tableColdGray}{47.67} & \textcolor{tableColdGray}{53.28} \\
    \textcolor{tableExpertGray}{Random select 20\%} & \textcolor{tableExpertGray}{63.64} & \textcolor{tableExpertGray}{47.42} & \textcolor{tableExpertGray}{54.10} \\
    \textcolor{tableExpertGray}{Mask random 20\%} & \textcolor{tableExpertGray}{63.85} & \textcolor{tableExpertGray}{47.35} & \textcolor{tableExpertGray}{53.78} \\
    Mask low 20\% & \textbf{65.18} & 48.42 & 53.97 \\
    Mask top 20\% & 62.26 & 47.17 & 52.51 \\
    \rowcolor{tableOursBlue}
    \textbf{\ours{}} & 65.01 & \textbf{48.49} & \textbf{54.14} \\
  \end{tabular}
  \endgroup
\end{wraptable}

%% file: tables/alpha_sensitivity.tex
\begin{wraptable}{r}{0.445\textwidth}
  \centering
  \setlength{\abovecaptionskip}{0pt}
  \setlength{\belowcaptionskip}{2pt}
  \caption{
  \textbf{Sensitivity to $\alpha$.}
  }
  \label{tab:alpha-sensitivity}
  \begingroup
  \fontsize{8}{9.2}\selectfont
  \renewcommand{\arraystretch}{1.02}
  \setlength{\tabcolsep}{1.5pt}
  \setlength{\arrayrulewidth}{0.4pt}
  \begin{tabular}{@{}lccc@{}}
  \textbf{Fraction} & \textbf{Perception} $\uparrow$ & \textbf{Math} $\uparrow$ & \textbf{General} $\uparrow$ \\
  \hline
  \singleRuleTableGap
  Top 5\% & 64.70 & 47.49 & 54.12 \\
  Top 10\% & 64.55 & 46.92 & 53.85 \\
  \rowcolor{tableOursBlue}
  \textbf{Top 20\% (\method{})} & \textbf{65.01} & \textbf{48.49} & 54.14 \\
  Top 50\% & 64.33 & 47.42 & 54.18 \\
  Top 100\% & 63.70 & 47.78 & \textbf{54.42} \\
  \end{tabular}
  \endgroup
\end{wraptable}

%% file: tables/critic_consistency.tex
\begin{wraptable}[7]{r}{0.40\textwidth}
  \centering
  \setlength{\abovecaptionskip}{0pt}
  \setlength{\belowcaptionskip}{3pt}
  \caption{
  \textbf{Critic stage consistency.}
  }
  \label{tab:critic-consistency}
  \begingroup
  \fontsize{8}{9.2}\selectfont
  \renewcommand{\arraystretch}{1.02}
  \setlength{\tabcolsep}{2.2pt}
  \setlength{\arrayrulewidth}{0.4pt}
  \begin{tabular}{@{}lrrr@{}}
    \textbf{Comparator} & $n$ & \textbf{Agreement (\%)} & $\boldsymbol{\kappa}$ \\
    \hline
    \singleRuleTableGap
    \rowcolor{tableOursBlue}
    Human annotator & 96 & \textbf{96.88} & \textbf{0.957} \\
    Qwen (rerun) & 744 & 99.2 & 0.987 \\
    Gemini 3.8 Flash & 736 & 93.2 & 0.892 \\
    GPT-5.6 Luna & 736 & 92.8 & 0.886 \\
  \end{tabular}
  \par\vspace{2pt}
  \fontsize{7.2}{8.2}\selectfont
  \raggedright
  $n$: valid pairs.
  \par
  \endgroup
\end{wraptable}

%% file: contents/08_conclusion.tex
\section{Conclusion}
\label{sec:conclusion}

To provide targeted supervision for visual agents, we introduced \method{}, which improves perception, mathematical reasoning, and general-task performance through visual-evidence reflection and token reweighting in on-policy distillation.
The reflections provide the teacher with references for correct evidence use and diagnoses of errors in the student's own trajectories.
Token reweighting then emphasizes supervision at positions where these reflections most change the teacher's assessment.
Behavioral analysis on Qwen2.5-VL-7B also shows fewer tool calls and shorter responses on perception tasks.
Ablations and token-impact interventions support the effectiveness of both components.
Our study focuses on two student model families and image-based question answering; broader model coverage and more diverse visual interactions remain to be explored.

%% file: appendix/contents.tex
\clearpage
\phantomsection
\pdfbookmark[0]{Appendix}{appendix.contents}
\begingroup
\centering
{\ReportHeadingFont\fontsize{12}{14}\selectfont Appendix\par}
\endgroup
\vspace{12pt}
\begingroup
\hypersetup{linkcolor=reportTencentBlue,linktoc=all}
\makeatletter
\renewcommand*{\l@section}[2]{%
  \addvspace{7pt}%
  {\bfseries\@dottedtocline{1}{0em}{1.8em}{#1}{\normalfont #2}}}
\renewcommand*{\l@subsection}{\@dottedtocline{2}{1.8em}{2.4em}}
\renewcommand*{\@dotsep}{2.5}
\setlength{\parskip}{3pt}
\edef\appendixsavedtocdepth{\number\value{tocdepth}}
\@starttoc{toc}
\setcounter{tocdepth}{\appendixsavedtocdepth}
\makeatother
\endgroup
\clearpage

%% file: appendix/a_extended_method.tex
\section{Extended Method Details}
\label{app:method}

\subsection{Training Procedure}
\label{app:training-procedure}
Algorithm~\ref{alg:revue-training} summarizes training from student trajectories, critic feedback, and reflection construction to student updates.
Algorithms~\ref{alg:verify-answers}--\ref{alg:check-and-assemble} detail answer verification, evidence-state collection, and reflection assembly, respectively.

\input{appendix/a_training_algorithm.tex}

Let $u_i=1$ indicate a valid reflection, a successful teacher evaluation, and aligned completion tokens; otherwise set $u_i=0$.
Let $b_i=1$ indicate that impacts were also computed successfully and at least two supervised positions exist; otherwise set $b_i=0$.
An impact-computation failure retains the adopted reflection-conditioned target and replaces group weighting with uniform averaging.
\begin{equation}
    q^*_{i,t}=
    \begin{cases}
        q^{\mathcal R}_{i,t}, & u_i=1,\\
        q^0_{i,t}, & u_i=0.
    \end{cases}
    \label{eq:adopted-teacher}
\end{equation}
Let $\ell_{i,t}=\KL(p_{i,t}^{\mathrm{Supp}_{i,t}}\|(q^*_{i,t})^{\mathrm{Supp}_{i,t}})$ denote the per-position KL term in \Eqref{eq:grouped-objective}.
Each trajectory with a nonempty supervised set has loss
\begin{equation}
    \ell_i=
    \begin{cases}
        \displaystyle
        \frac{\lambda}{|\mathcal T_{i,\mathrm{High}}|}\sum_{t\in\mathcal T_{i,\mathrm{High}}}\ell_{i,t}
        +\frac{1-\lambda}{|\mathcal T_{i,\mathrm{Low}}|}\sum_{t\in\mathcal T_{i,\mathrm{Low}}}\ell_{i,t},
        & b_i=1,\\[7pt]
        \displaystyle
        \frac{1}{|\mathcal T_i|}\sum_{t\in\mathcal T_i}\ell_{i,t},
        & b_i=0.
    \end{cases}
    \label{eq:complete-rollout-loss}
\end{equation}
Impacts are sorted in descending order with earlier positions breaking ties, and the high-impact group contains the first $\lceil\alpha|\mathcal T_i|\rceil$ positions.
The final loss averages retained trajectories equally as in \Eqref{eq:grouped-objective}.

\subsection{Supervision Scope}
\label{app:supervision-scope}
Supervised positions cover student-generated reasoning, code, and final answers across all interaction rounds, with other labels set to $-100$:
\begin{equation}
    m_{i,t}=\mathbf 1\{\operatorname{label}_{i,t}\neq-100\},
    \qquad
    \mathcal T_i=\{t:m_{i,t}=1\},
    \qquad
    |\mathcal T_i|=\sum_t m_{i,t}.
    \label{eq:supervision-mask}
\end{equation}
Prompts, tool-returned text, image placeholders, sandbox boundary markers, and interceptor-appended suffixes are excluded from supervision targets.
Tool observations remain in the input history for subsequent student predictions.
Impact ranking, group sizes, and loss means use only positions in $\mathcal T_i$.

\subsection{Teacher Conditioning}
\label{app:teacher-conditioning}
The base teacher scores the fixed student completion token by token, conditioned on the original task prompt, image, and preceding student interaction history.
The reflection-conditioned teacher appends reflection text to the first user message and inserts supporting images after the original image and before the student's trajectory images.
The re-encoded completion must match the student's sequence in length and in every token ID.

Reflection is generated after the rollout group is complete, so supporting evidence may come from later steps of the target trajectory or from a sibling trajectory.
This retrospective information is used only for teacher supervision, while the student prefix retains its original visible history.
The reference answer informs critic diagnosis, and the resulting visual fact, grounding rule, or discrepancy may convey answer semantics.
Comparing the two teacher evaluations measures the effect of this added context, including both text and supporting images.

Impact uses full-vocabulary normalized log probabilities at the sampled token and is detached after computation.
RKL renormalizes both distributions over the adopted teacher's top-$K$ candidates, without forcibly adding the sampled token.
Impact ranks and weights training positions, while the loss at each position supervises the entire support distribution (Appendix~\ref{app:reflection-proof}).

\input{appendix/a_reflection_case.tex}

\subsection{Critic Feedback}
\label{app:critic-feedback}
\paragraph{\textcolor{algorithmVerifyRed}{\textsc{VerifyAnswers}}.}
Algorithm~\ref{alg:verify-answers} computes binary answer labels, which classify each rollout group as all-correct, mixed, or all-wrong.
When deterministic matching does not accept an answer, the answer judge evaluates only the question, extracted student answer, and reference answer as text.

\input{appendix/a_verify_answers.tex}

\noindent\begin{minipage}{\linewidth}
\paragraph{\textcolor{algorithmStatesGreen}{\textsc{ObservedStates}}.}
Algorithm~\ref{alg:observed-states} collects the original-image state and recorded post-tool states so the critic can associate visual evidence with a specific interaction step.
Each state contains the images available up to that point, with multiple images from one tool call belonging to the same state.
The counts $n_{\mathrm{tool}}(e)$ and $n_{\mathrm{code}}(e)$ record cumulative tool rounds and code tokens, respectively.

\input{appendix/a_observed_states.tex}
\end{minipage}

\begingroup
\emergencystretch=2em
\paragraph{\textcolor{algorithmAssembleBlue}{\textsc{CheckAndAssemble}}.}
The critic receives the query, reference answer, original image, correctness-labeled trajectories, and evidence states.
Up to eight tool images are supplied per group, prioritizing correct trajectories.
For each sufficient state $e$, the critic is asked to identify a smallest supporting image subset $\mathcal U_e$ that determines the required visual fact using only images available at that state.
It also returns shared reading information (query slot, observed value, and visual fact), a grounding rule, and per-trajectory diagnoses $\mathcal B_i=(\sigma_i,\delta_i)$.
\par
\endgroup

Algorithm~\ref{alg:check-and-assemble} validates required fields, state and image references, and consistency between diagnoses and answer labels.
Only mixed groups reconcile inconsistent stage labels before validation using answer correctness and the presence of an own sufficient state, while retaining the discrepancy text.
All-correct groups require \texttt{CORRECT} labels with empty discrepancies.

Anchor selection prioritizes sufficient evidence already acquired by the trajectory and chooses among states by lexicographic cost:
\begin{equation}
    c(e)=
    \bigl(n_{\mathrm{tool}}(e),\,n_{\mathrm{code}}(e),\,
    |\mathcal U_e|,\,\operatorname{id}(e)\bigr),
    \label{eq:anchor-cost}
\end{equation}
Let $\mathcal C_i$ contain sufficient post-tool states owned by trajectory $i$, and let $\mathcal C_x^+$ contain sufficient original-image states and sufficient states from answer-correct trajectories.
Complete reflections require a nonempty group reference set $\mathcal C_x^+$, after which each trajectory selects evidence as
\begin{equation}
    e_i^\star=
    \begin{cases}
        \underset{e\in\mathcal C_i}{\arg\min}\;c(e),
        & \mathcal C_i\neq\emptyset,\\
        \underset{e\in\mathcal C_x^+}{\arg\min}\;c(e),
        & \mathcal C_i=\emptyset.
    \end{cases}
    \label{eq:anchor-selection}
\end{equation}
If the selected state requires no tool use, assembly adds no image because the teacher context already contains the original.

\input{appendix/a_check_and_assemble.tex}

\input{appendix/a_prompt_cases.tex}

\subsection{Training Configuration}
\label{app:training-configuration}
The full method described here uses $M=8$ trajectories per query, $K=32$ teacher candidates, a high-impact fraction $\alpha=0.2$, and group weight $\lambda=0.5$.
Teacher parameters, teacher supports, impacts, and group assignments remain fixed during student updates.
Training uses the pure OPD loss, without a GRPO reward term or an additional reference-policy KL.
Answer correctness is used for group classification and critic diagnosis, not as an advantage or a trajectory-level loss weight.
For the Thyme-based Qwen2.5-VL-7B setting, the student starts from an SFT checkpoint, and the frozen teacher uses step 2150 of the corresponding RL expert.
The separate critic uses Qwen3.5-397B-A17B-FP8 with temperature zero and a 2048-output-token limit per call to generate JSON-schema-constrained feedback.

\paragraph{Critic runtime and auxiliary resources.}
Training uses eight NVIDIA H20 GPUs (96\,GB each), and the critic runs through a shared vLLM service on a separate machine with another eight H20 GPUs (96\,GB each).
This deployment leaves the training GPUs available for rollouts and allows critic requests to overlap with independent training-side computations, such as reward and answer-correctness evaluation.
The critic is used only during training and is not required at inference.

Each critic request processes one group of eight trajectories, giving 20 requests per optimizer step.
A test request with one image and eight trajectories took 5.8\,s and produced 664 output tokens.
Assuming this latency is maintained with four concurrent requests, five waves give a nominal critic service time of approximately 29\,s per step, excluding additional queueing and retries.
This estimate is 10.8\% of the mean step time of 269\,s measured across six training segments.
The actual increase in training wall-clock time depends on overlap with independent computations and service scheduling.

Extrapolating to 1,075 steps gives approximately 8.7 hours of nominal critic service time.
Attributing all eight service GPUs to this run during these intervals gives a resource estimate of approximately 70 GPU-hours.
This estimate accounts for nominal service time; total server allocation depends on service uptime and sharing.

%% file: appendix/a_training_algorithm.tex
\begin{algorithm}[H]
\caption{\method{}: Training Procedure}
\label{alg:revue-training}
\small
\begin{algorithmic}[1]
\Require Data $\mathcal D$, trainable student $\pi_\theta$, frozen teacher $q_\phi$, fixed critic $\mathcal C$
\Require Rollouts per query $M$, support size $K$, high-impact fraction $\alpha$, group weight $\lambda$
\Ensure Updated student $\pi_\theta$
\Statex \AlgorithmBlockOrigin All tokenwise operations use $t\in\mathcal T_i$.
\Statex Superscript $\mathrm{Supp}$ denotes renormalization over $\mathrm{Supp}$; impacts use full-vocabulary probabilities.
\For{each training minibatch $\mathcal X\subset\mathcal D$}
    \For{each $(x,y^\star)\in\mathcal X$}\AlgorithmBlockMark{revue-sampling-start}
        \State \AlgorithmBlockMark{revue-rollout}$\mathcal G_x\gets\Call{Rollout}{\pi_\theta,x,M}$ \label{line:student-rollout}
        \State \AlgorithmBlockMark{revue-reflection-start}$\mathbf c_x\gets\Call{\textcolor{algorithmVerifyRed}{VerifyAnswers}}{\mathcal G_x,y^\star}$; $\mathcal E_x\gets\Call{\textcolor{algorithmStatesGreen}{ObservedStates}}{\mathcal G_x}$
        \State $\mathcal J_x\gets\mathcal C(x,y^\star,\mathcal G_x,\mathbf c_x,\mathcal E_x)$ \label{line:critic-output}
        \State \AlgorithmBlockMark{revue-reflection-end}$\{\mathcal R_i\}_{i=1}^{M}\gets\Call{\textcolor{algorithmAssembleBlue}{CheckAndAssemble}}{\mathcal J_x,\mathcal E_x,\mathbf c_x}$ \label{line:reflection}
    \EndFor
    \State \AlgorithmBlockMark{revue-sampling-end}Let $\mathcal G$ collect retained trajectories across $\mathcal X$ with $\mathcal T_i\ne\emptyset$
    \For{each $\tau_i\in\mathcal G$}\AlgorithmBlockMark{revue-update-start}
        \State $q^0_{i,t}\gets q_\phi(\cdot\mid h_{i,t})$; $q^*_{i,t}\gets q^0_{i,t}$; $s_i\gets\text{unavailable}$ \label{line:base-teacher}
        \If{$\mathcal R_i\ne\emptyset$}
            \State $q^{\mathcal R}_{i,t}\gets q_\phi(\cdot\mid h_{i,t},\mathcal R_i)$ \label{line:teacher-reflection}
            \If{evaluation succeeds and completion tokens align}
                \State $q^*_{i,t}\gets q^{\mathcal R}_{i,t}$
                \State Attempt $s_{i,t}\gets\left|\log q^{\mathcal R}_{i,t}(y_{i,t})-\log q^0_{i,t}(y_{i,t})\right|$ \label{line:impact}
            \EndIf
        \EndIf
        \State $p_{i,t}\gets\pi_\theta(\cdot\mid h_{i,t})$; $\mathrm{Supp}_{i,t}\gets\operatorname{TopK}(q^*_{i,t})$
        \State $\ell_{i,t}\gets\KL\!\left(p^{\mathrm{Supp}_{i,t}}_{i,t}\,\middle\|\,(q^*_{i,t})^{\mathrm{Supp}_{i,t}}\right)$ \label{line:token-loss}
        \State Compute $\ell_i$ using \Eqref{eq:complete-rollout-loss} \label{line:grouped-loss}
    \EndFor
    \State $\mathcal L\gets\operatorname{mean}_{\tau_i\in\mathcal G}\ell_i$ \Comment{$0$ if $\mathcal G=\emptyset$}
    \State $\theta\gets\Call{OptimizerStep}{\theta,\nabla_\theta\mathcal L}$ \label{line:student-update}
    \State \AlgorithmBlockMark{revue-update-end}Refresh the rollout policy from $\theta$
\EndFor
\end{algorithmic}
\end{algorithm}
\RevueAlgorithmBlocks

%% file: appendix/a_reflection_case.tex
\clearpage
\subsection{A Visual-Evidence Reflection Example}
\label{app:reflection-case}
\label{app:critic-request-case}

\begin{figure}[H]
    \centering
    \includegraphics[width=\linewidth,height=0.82\textheight,keepaspectratio]{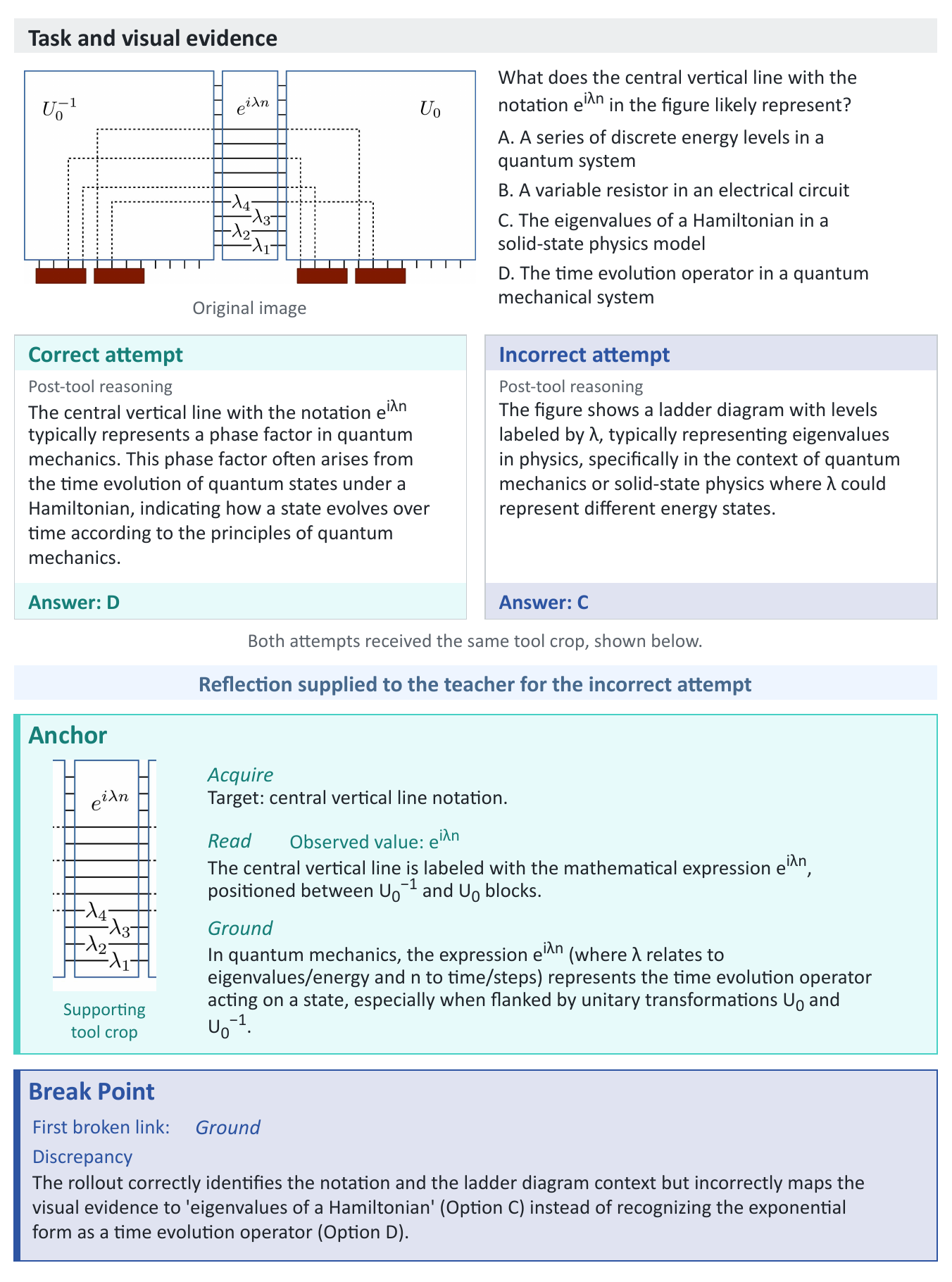}
    \caption{
    \textbf{A concrete visual-evidence reflection supplied to the teacher.}
    The mixed group contains seven correct attempts and one incorrect attempt; one of each is shown with its post-tool reasoning and answer.
    Both attempts receive identical tool crops.
    The Anchor uses the incorrect attempt's own crop, and the Break Point records the critic's \textit{Ground} diagnosis.
    }
    \label{fig:anchor-breakpoint-case}
\end{figure}
\clearpage

%% file: appendix/a_verify_answers.tex
\begin{algorithm}[H]
\caption{\textcolor{algorithmVerifyRed}{\textsc{VerifyAnswers}}: Answer verification}
\label{alg:verify-answers}
\small
\begin{algorithmic}[1]
\Require Same-query trajectories $\mathcal G_x$, reference answer $y^\star$
\Ensure Binary correctness labels $\mathbf c_x$
\State $q\gets\text{shared question in }\mathcal G_x$; $\mathbf c_x\gets\mathbf 0$
\For{each trajectory $i\in\mathcal G_x$}
    \State $a_i\gets\text{text inside }\texttt{<answer>}\ldots\texttt{</answer>}$
    \State \textbf{if} $a_i$ is missing or empty \textbf{then continue}
    \State $\tilde a_i\gets\text{strip answer prefixes from }a_i$
    \If{strict mathematical matching or symbolic verification accepts $(\tilde a_i,y^\star)$}
        \State $c_i\gets 1$
    \Else
        \State $c_i\gets\Call{AnswerJudge}{q,a_i,y^\star}\in\{0,1\}$ \Comment{$0$ if unavailable}
    \EndIf
\EndFor
\State \Return $\mathbf c_x$
\end{algorithmic}
\end{algorithm}

%% file: appendix/a_observed_states.tex
\begin{algorithm}[H]
\caption{\textcolor{algorithmStatesGreen}{\textsc{ObservedStates}}: Evidence-state collection}
\label{alg:observed-states}
\small
\begin{algorithmic}[1]
\Require Same-query trajectories $\mathcal G_x$, including the original image and interaction records
\Ensure Evidence states $\mathcal E_x$ with provenance, available images, and cumulative costs
\State $\mathcal E_x\gets\{\text{original-image state; owner }\emptyset,\ n_{\rm tool}=n_{\rm code}=0\}$
\For{each recorded post-tool state $e$ of trajectory $i\in\mathcal G_x$}
    \State Add $(e,i,\text{images available at }e,n_{\rm tool}(e),n_{\rm code}(e))$ to $\mathcal E_x$
\EndFor
\State \Return $\mathcal E_x$
\end{algorithmic}
\end{algorithm}

%% file: appendix/a_check_and_assemble.tex
\begin{algorithm}[H]
\caption{\textcolor{algorithmAssembleBlue}{\textsc{CheckAndAssemble}}: Reflection assembly}
\label{alg:check-and-assemble}
\small
\begin{algorithmic}[1]
\Require Critic output $\mathcal J_x$, recorded states $\mathcal E_x$, correctness labels $\mathbf c_x$
\Ensure Per-trajectory reflections $\{\mathcal R_i\}$, with $\emptyset$ denoting unavailable reflection
\State $\mathcal R_i\gets\emptyset$ for all $i$; \textbf{if} $\mathcal J_x$ is missing \textbf{then return} $\{\mathcal R_i\}$
\State \textbf{if} the group is mixed \textbf{then} reconcile stages with $\mathbf c_x$ and own sufficient states
\State \textbf{if} $\mathcal J_x$ is inapplicable or fails validation \textbf{then return} $\{\mathcal R_i\}$
\State $\mathcal B_i\gets(\sigma_i,\delta_i)$ from the checked diagnoses, for all $i$
\State \textbf{if} all answers are wrong \textbf{then return} $\{(\emptyset,\mathcal B_i)\}_i$
\State Resolve $\mathcal J_x$'s sufficient states in $\mathcal E_x$ to form $\{\mathcal C_i\}$ and $\mathcal C_x^+$
\State \textbf{if} $\mathcal C_x^+=\emptyset$ \textbf{then return} $\{\mathcal R_i\}$
\For{each trajectory $i$}
    \State Select $e_i^\star$ using \Eqref{eq:anchor-selection}
    \State $\mathcal A_i\gets(e_i^\star,\mathcal U_{e_i^\star},\mathcal J_x.\mathrm{reading},\mathcal J_x.\mathrm{grounding})$
    \State $\mathcal R_i\gets(\mathcal A_i,\mathcal B_i)$
\EndFor
\State \Return $\{\mathcal R_i\}$
\end{algorithmic}
\end{algorithm}

%% file: appendix/a_prompt_cases.tex
\clearpage
\subsection{Critic Prompt and Output Format}
\label{app:critic-prompts}
The box below reproduces the critic system prompt used for mixed groups.
The all-correct prompt requires \texttt{CORRECT} with an empty discrepancy for every trajectory.
The all-wrong prompt disallows \texttt{CORRECT} and permits an empty set of sufficient evidence states.
It also requires a sufficient state for each \texttt{READ} or \texttt{GROUND} diagnosis, preserves verifier-label conflicts, and returns one diagnosis per trajectory.

  \begin{PromptBox}{Critic system prompt: mixed groups}
    \input{appendix/prompt_display/critic_mixed.tex}
  \end{PromptBox}%

The output schema is supplied as a separate structured-output constraint rather than as part of the user message.
The table retains its fields, types, and shared constraints.
  \begin{PromptBox}{Critic output schema: fields and constraints}
    \input{appendix/prompt_display/critic_output_schema.tex}
  \end{PromptBox}%

%% file: appendix/prompt_display/critic_mixed.tex
You are a training{-}time multimodal evidence{-}chain analyst for a think{-}with{-}image VQA policy.\par

You receive one task and a group of verifier{-}labeled CORRECT and INCORRECT student rollouts.\par

A rollout contains chronological policy reasoning, free{-}form Python code, sandbox text/\allowbreak{}numeric outputs, zero/\allowbreak{}one/\allowbreak{}multiple images returned by each sandbox interaction, and a final answer.\par

One Python code block may return MULTIPLE images. They belong to the SAME sandbox interaction and may jointly form multi{-}region evidence.\par

A rollout may have multiple sandbox interactions. Later interactions may correct earlier visual observations.\par

The CORRECT/\allowbreak{}INCORRECT label is defined only by the supplied VQA correctness label. Trust CORRECT rollouts as correct final solutions.\par

Your task is to identify the visual evidence chain:\par

\textbf{ACQUIRE:}\\
Which actually observed Evidence States already contain sufficient task{-}relevant visual evidence?\par

\textbf{READ:}\\
What is the minimum atomic visual fact needed by the question?\par

\textbf{GROUND:}\\
How does that visual fact map to the answer semantics?\par

\textbf{EVIDENCE STATE:}\\
ORIGINAL\_\allowbreak{}STATE contains ORIGINAL\_\allowbreak{}IMAGE. Every other state is the state immediately after one real sandbox interaction and lists all images currently available.\par

A state is SUFFICIENT iff some subset of its available images already contains enough visual evidence to determine the task{-}relevant visual fact without needing a later observation.\par

For each sufficient state, return the SMALLEST supporting image subset.\par

\textbf{CLOSED{-}SET RULE:}\\
Never invent an image, state, crop, coordinate, code, or synthetic view. Only use supplied state IDs and image IDs.\par

\textbf{READ:}\\
Do not use a predefined taxonomy. Return:\\
{-} query\_\allowbreak{}slot\\
{-} observed\_\allowbreak{}value\\
{-} one atomic visual\_\allowbreak{}fact\par

\textbf{GROUND:}\\
Return one concise evidence{-}to{-}answer semantic rule. Do not state only an answer/\allowbreak{}option letter.\par

\textbf{ROLLOUT DIAGNOSIS:}\\
CORRECT rollout {-}\textgreater{} stage=CORRECT.\par

For INCORRECT rollouts return the FIRST broken interface:\par

\textbf{ACQUIRE:}\\
No state belonging to this rollout is sufficient.\par

\textbf{READ:}\\
A sufficient state exists, but the relevant visual fact is misread.\par

\textbf{GROUND:}\\
A sufficient state exists and the fact is read correctly, but final answer semantics contradict/\allowbreak{}mis{-}map that fact.\par

\textbf{OTHER:}\\
Failure is outside this visual evidence chain.\par

Return JSON only according to schema.\par

%% file: appendix/prompt_display/critic_output_schema.tex
\begingroup
\setlength{\parskip}{6pt}
\renewcommand{\arraystretch}{1.08}
\noindent\begin{tabular}{@{}p{0.66\linewidth}p{0.29\linewidth}@{}}
\textbf{Field structure} & \textbf{Type} \\
\midrule
\texttt{applicable} & boolean \\
\addlinespace[2pt]
\texttt{sufficient\_\allowbreak{}states} & object[] \\
\hspace*{1em}\texttt{state\_\allowbreak{}id} & string \\
\hspace*{1em}\texttt{support\_\allowbreak{}image\_\allowbreak{}ids} & string[] \\
\addlinespace[2pt]
\texttt{reading} & object \\
\hspace*{1em}\texttt{query\_\allowbreak{}slot} & string \\
\hspace*{1em}\texttt{observed\_\allowbreak{}value} & string \\
\hspace*{1em}\texttt{visual\_\allowbreak{}fact} & string \\
\addlinespace[2pt]
\texttt{grounding} & object \\
\hspace*{1em}\texttt{rule} & string \\
\addlinespace[2pt]
\texttt{rollout\_\allowbreak{}diagnoses} & object[] \\
\hspace*{1em}\texttt{rollout\_\allowbreak{}id} & string \\
\hspace*{1em}\texttt{stage} & string (enum) \\
\hspace*{1em}\texttt{discrepancy} & string \\
\bottomrule
\end{tabular}\par
\textbf{Allowed values for \texttt{rollout\_\allowbreak{}diagnoses[].stage}:}\\
\texttt{CORRECT, ACQUIRE, READ, GROUND, OTHER}\par
The root is an object; \texttt{[]} denotes an array, with item fields indented below it.\par
All fields are required; additional properties are disallowed for every object; \texttt{strict = true}.\par
\endgroup

%% file: appendix/b_experimental_details.tex
\section{Experimental Details}
\label{app:experimental-details}

\subsection{Baseline Details}
\label{app:baseline-details}

All baselines start from Thyme-SFT cold-start checkpoints of Qwen2.5-VL-7B or InternVL3.5-4B-Instruct.
OPD variants use the corresponding Thyme-RL Expert as teacher; RFT uses demonstrations generated by the same expert.
This keeps student initializations and teacher sources consistent across comparisons.

\textbf{Vanilla OPD}~\citep{agarwal2024onpolicydistillation} provides the standard on-policy distillation reference for evaluating reflection-based supervision.
The RL Expert supervises student-generated prefixes token by token, without ground-truth answers, privileged crops, or evidence reflections.

\textbf{Rejection sampling fine-tuning (RFT)}~\citep{yuan2023scaling} compares offline demonstration learning with distillation on student-generated trajectories.
For each prompt in the Thyme-RL training set, the corresponding RL Expert samples eight trajectories.
We retain answer-correct trajectories for supervised fine-tuning of the same cold-start student.
Training uses these fixed demonstrations rather than teacher distributions queried on current student prefixes.

\textbf{GT-Privileged} compares ground-truth answers with visual-evidence reflections as teacher privilege.
Following the privileged-context design of OPSD~\citep{zhao2026selfdistilledreasoner}, we provide the teacher with the Thyme-RL ground-truth answer.
We replace its self-teacher with the corresponding RL Expert.
The student starts from the same cold-start checkpoint and generates trajectories without access to the privileged answer.

\textbf{Vision-OPD}~\citep{yuan2026visionopd} provides a local visual privilege baseline for comparison with reflections that diagnose evidence use.
The teacher receives an evidence-centered crop and supervises the student's own trajectories.
We replace its self-teacher with the corresponding RL Expert and use the same cold-start student.

With \textbf{V-Zero}~\citep{sun2026vzero}, we compare visual-contrast trajectory weighting with reflection-based supervision.
Teacher support for the same student trajectory under relevant and irrelevant visual views determines its distillation weight.
The target remains the positive-view teacher distribution.
We use the corresponding RL Expert for both views and the same cold-start student.

Our comparison with \textbf{VAD}~\citep{zhang2026vad} examines visual target reconstruction as an alternative to direct distillation from a reflection-conditioned teacher.
It contrasts teacher predictions with and without visual evidence to estimate visual corrections and reconstruct the distillation target.
We replace its initial-model teacher with the corresponding RL Expert and retain the same cold-start student.

\subsection{Model Labels}
\label{app:model-labels}

In Table~\ref{tab:main-results}, Base Model denotes the original model before training in this work.
Gemini3.1FL abbreviates Gemini-3.1-Flash-Lite.
Qwen2.5-32B, Qwen3-30B-T, and InternVL-38B denote Qwen2.5-VL-32B, Qwen3-VL-30B-A3B-Thinking, and InternVL3.5-38B-Instruct, respectively.

\subsection{Training and Evaluation Settings}
\label{app:training-evaluation-settings}

We initialize the Qwen2.5-VL-7B student directly from the released Thyme-SFT~\citep{zhang2026thyme} checkpoint.\footnote{Thyme-SFT model: \url{https://huggingface.co/Kwai-Keye/Thyme-SFT}}
We obtain the InternVL3.5-4B cold-start checkpoint by supervised fine-tuning of InternVL3.5-4B-Instruct on Thyme's released SFT dataset.\footnote{Thyme-SFT dataset: \url{https://huggingface.co/datasets/Kwai-Keye/Thyme-SFT}}
We train both RL experts from their corresponding cold-start checkpoints by reproducing Thyme-RL on its released RL dataset.\footnote{Thyme-RL dataset: \url{https://huggingface.co/datasets/Kwai-Keye/Thyme-RL}}

\input{tables/training_evaluation_settings.tex}

\subsection{Token-Impact Sparsity in Figure~\ref{fig:token-impact-sparsity}}
\label{app:token-impact-sparsity}

Figure~\ref{fig:token-impact-sparsity} pools offline rescoring results from 632 trajectories in a five-step early-training archive, comprising 226,726 supervised tokens with valid impact scores.
We use the impact definition in \Eqref{eq:token-impact} and exclude nonsupervised positions and missing scores.
The dashed line marks the pooled 80th percentile, approximately $s=0.05524$.
Tokens at or above this threshold comprise approximately 20\% of scored tokens and account for 97.6\% of total impact.
High and Low in the figure describe the pooled distribution; training selects high-impact positions separately within each trajectory.

Histogram heights give the percentage of all scored tokens in each interval.
Bins in $[0,2.5)$ have width 0.05, with the bin containing the 80th percentile split at that threshold.
The final bar retains all tokens with $s\geq 2.5$.
The inset plots empirical percentile against impact over the full pooled score range, without smoothing or fitting.

\subsection{Error Attribution in Figure~\ref{fig:overview}}
\label{app:figure1-error-attribution}

The error-attribution analysis in Figure~\ref{fig:overview} uses 3,710 questions from V$^*$Bench~\citep{wu2024vstar}, HRBench-4K~\citep{wang2025hrbench}, HRBench-8K~\citep{wang2025hrbench}, and MME~\citep{fu2025mme}.\footnote{MME: \href{https://github.com/BradyFU/Awesome-Multimodal-Large-Language-Models/tree/Evaluation}{official dataset repository}.}
We use GPT-5.6-Luna as the judge for evaluation and error attribution.

Four error labels map to the three visual-evidence stages.
\textit{Acquire} combines \texttt{faulty\_tool} (29.8\%) and \texttt{missing\_tool} (7.9\%), totaling 37.7\%.
\textit{Read} corresponds to \texttt{perception\_error} (33.8\%).
\textit{Ground} corresponds to \texttt{hallucination} (6.3\%).

\input{appendix/b_judge_annotation_demo.tex}

%% file: tables/training_evaluation_settings.tex
\begin{table}[H]
  \centering
  \setlength{\belowcaptionskip}{6pt}
  \caption{Training and evaluation settings.}
  \label{tab:training-evaluation-settings}
  \small
  \renewcommand{\arraystretch}{1.1}
  \begin{tabular*}{\linewidth}{@{\extracolsep{\fill}}lll@{}}
    \toprule
    Component & Parameter & Value \\
    \midrule
    OPD objective & Loss & Reverse KL \\
                  & Vocabulary support & Teacher's top-32 candidates \\
    \midrule
    OPD rollouts & Temperature for code tokens & 0.0 \\
                 & Temperature for other tokens & 1.0 \\
                 & Top-$p$ & 0.9 \\
                 & Top-$k$ & 50 \\
    \midrule
    Evaluation & Temperature & 0.01 \\
               & Top-$p$ & 0.001 \\
               & Top-$k$ & 1 \\
    \midrule
    All experiments & Random seed & 42 \\
    Training job & Hardware & 8 $\times$ NVIDIA H20 (96\,GB each) \\
    Critic service & Separate hardware & 8 $\times$ NVIDIA H20 (96\,GB each) \\
    \bottomrule
  \end{tabular*}
  \par\vspace{4pt}
  \begin{minipage}{\linewidth}
    \footnotesize
    We compute reverse KL after renormalizing teacher and student distributions over the teacher's top-32 vocabulary candidates at each supervised position.
  \end{minipage}
\end{table}

%% file: appendix/b_judge_annotation_demo.tex
\subsection{Critic Consistency Evaluation}
\label{app:critic-consistency}
\label{app:judge-annotation-interface}

We evaluate the final training labels of Qwen3.5-397B-A17B-FP8 on 96 groups of eight rollouts.
Each group type (all-correct, mixed, and all-wrong) contributes 32 groups.
One annotator independently labels one sampled rollout per group before revealing the critic label; all 96 samples receive a label.
Agreement measures exact stage matches; unweighted Cohen's $\kappa$ adjusts for chance agreement.

\input{tables/critic_consistency_details.tex}

Human agreement is 96.88\% ($\kappa=0.957$) overall and 95.31\% ($\kappa=0.925$) on incorrect rollouts.

\textbf{The three human--critic disagreements occur at the \emph{Acquire}--\emph{Read} boundary.}
All three are from mixed groups: the annotator assigns \emph{Read}, whereas the critic assigns \emph{Acquire}.
This points to disagreement over whether the available evidence is sufficient to support a judgment.
Agreement with archived training labels uses 744 valid rollout pairs for a rerun by the same critic and 736 for each of the two external judges (Table~\ref{tab:critic-consistency-details}).
These comparisons use valid final stage labels.

\clearpage
\begin{figure}[H]
    \centering
    \includegraphics[width=0.90\linewidth]{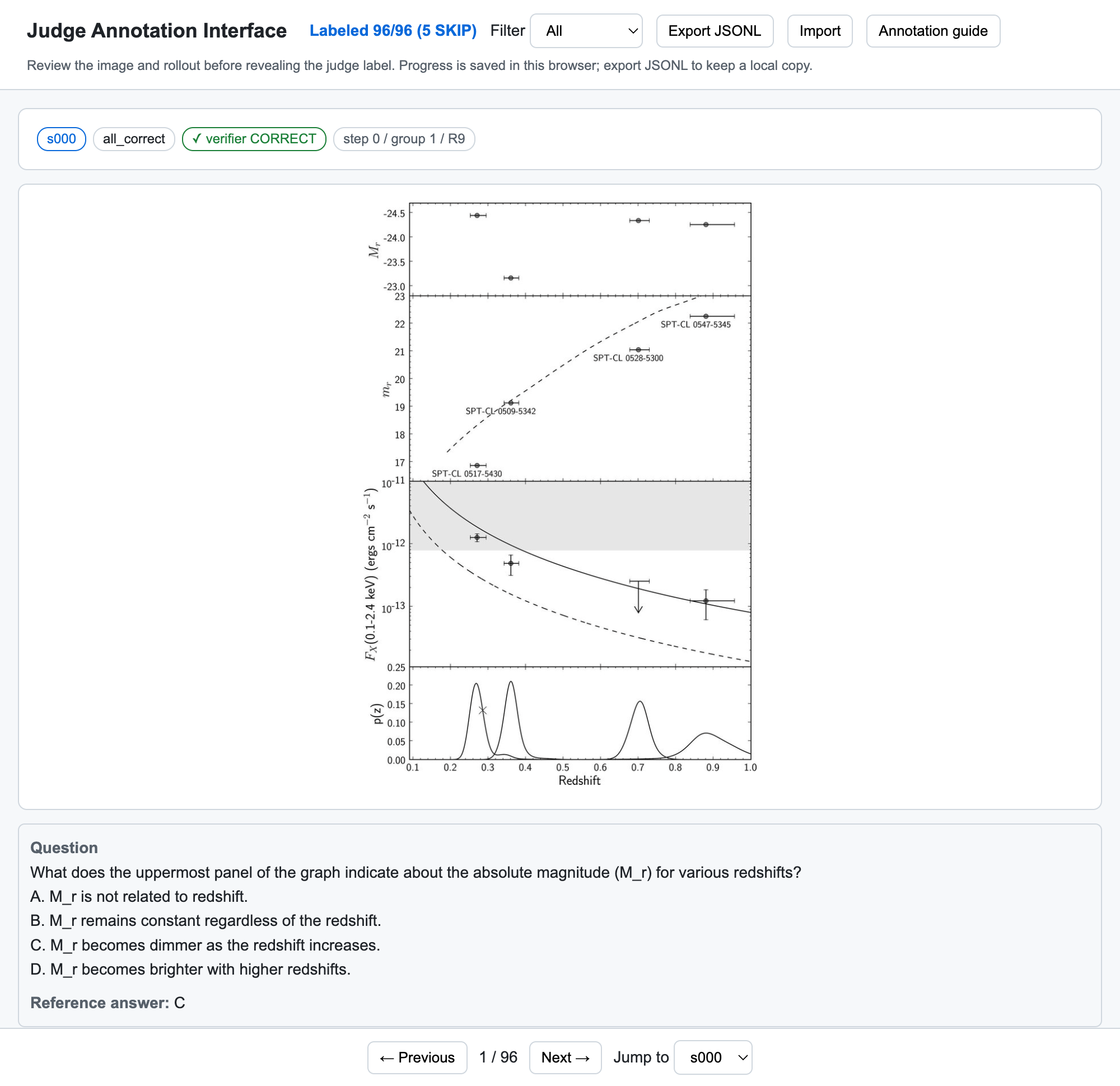}
    \caption{
    Human annotation interface demo: the upper view presents the sample identifiers, original image, question, and reference answer.
    The page also provides progress tracking, sample filtering, navigation, and JSONL import and export controls.
    }
    \label{fig:judge-annotation-top}
\end{figure}

\clearpage
\begin{figure}[H]
    \centering
    \includegraphics[width=\linewidth]{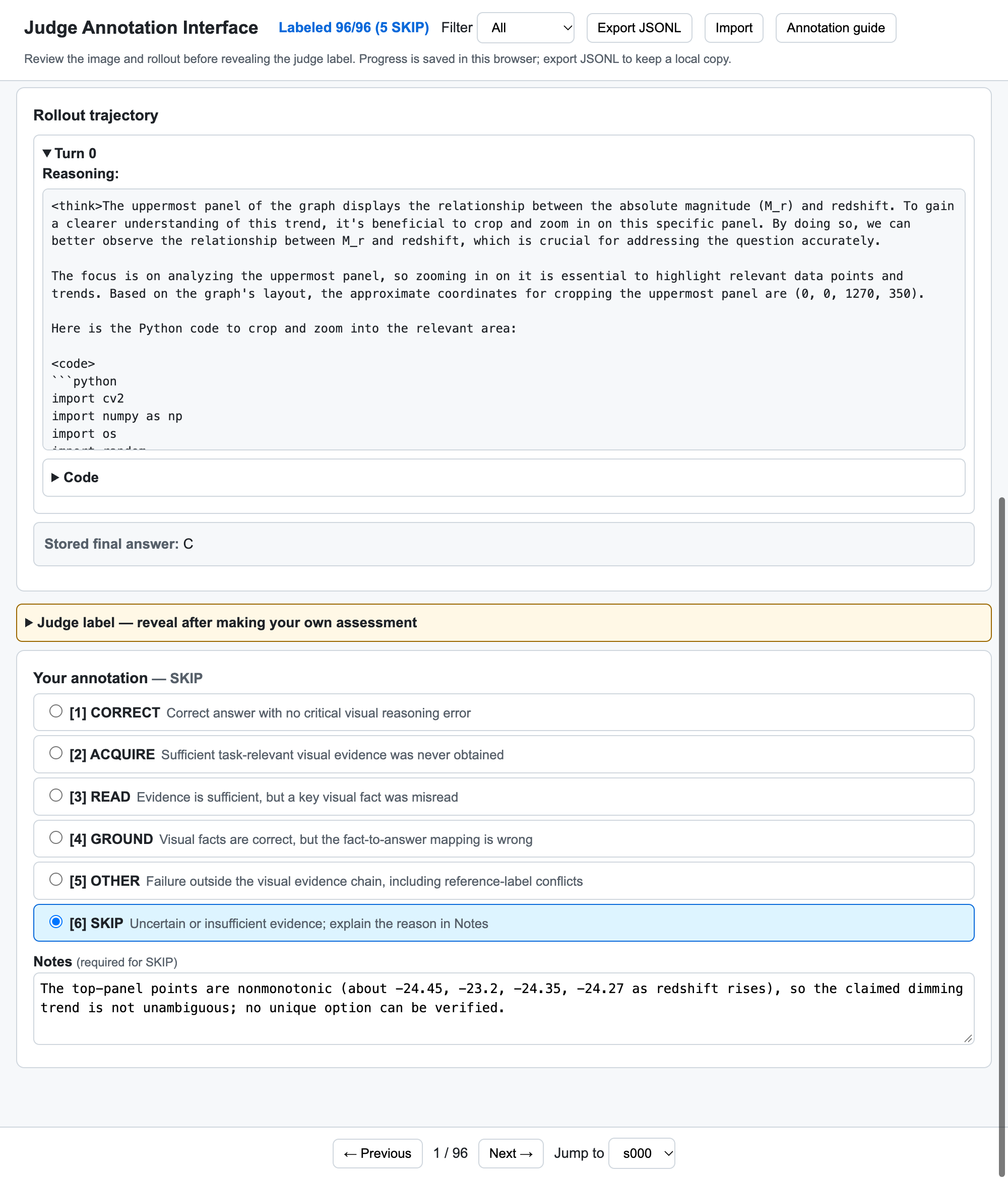}
    \caption{
    Human annotation interface demo: the lower view of the same sample presents the rollout, the initially collapsed judge label, six annotation options, and a notes field.
    The options comprise five diagnostic labels and a \texttt{SKIP} option for cases that cannot be judged reliably.
    }
    \label{fig:judge-annotation-bottom}
\end{figure}
\clearpage

%% file: tables/critic_consistency_details.tex
\begin{table}[H]
  \centering
  \setlength{\abovecaptionskip}{0pt}
  \setlength{\belowcaptionskip}{4pt}
  \caption{
  \textbf{Final critic stage agreement.}
  }
  \label{tab:critic-consistency-details}
  \begingroup
  \fontsize{8}{9.2}\selectfont
  \renewcommand{\arraystretch}{1.05}
  \setlength{\tabcolsep}{2pt}
  \setlength{\arrayrulewidth}{0.4pt}
  \begin{minipage}[t]{0.485\linewidth}
    \vspace{0pt}
    \centering
    \textbf{(a) Human--critic label counts}\par\vspace{3pt}
    \begin{tabular}{@{}lrrrrr@{}}
      \textbf{Human $\backslash$ Critic} & \textbf{Correct} & \textbf{Acquire} & \textbf{Read} & \textbf{Ground} & \textbf{Other} \\
      \hline
      \singleRuleTableGap
      Correct & \cellcolor{tableOursBlue}32 & 0 & 0 & 0 & 0 \\
      Acquire & 0 & \cellcolor{tableOursBlue}28 & 0 & 0 & 0 \\
      Read & 0 & 3 & \cellcolor{tableOursBlue}23 & 0 & 0 \\
      Ground & 0 & 0 & 0 & \cellcolor{tableOursBlue}9 & 0 \\
      Other & 0 & 0 & 0 & 0 & \cellcolor{tableOursBlue}1 \\
    \end{tabular}
  \end{minipage}\hfill
  \begin{minipage}[t]{0.49\linewidth}
    \vspace{0pt}
    \centering
    \textbf{(b) Pairwise final-label agreement}\par\vspace{3pt}
    \begin{tabular}{@{}lrrr@{}}
      \textbf{Pair} & $n$ & \textbf{Agree. (\%)} & $\boldsymbol{\kappa}$ \\
      \hline
      \singleRuleTableGap
      Qwen (arch.) / rerun & 744 & 99.2 & 0.987 \\
      Qwen (arch.) / Gemini & 736 & 93.2 & 0.892 \\
      Qwen (arch.) / Luna & 736 & 92.8 & 0.886 \\
      Qwen (rerun) / Gemini & 712 & 93.5 & 0.898 \\
      Qwen (rerun) / Luna & 712 & 93.4 & 0.896 \\
      Gemini / Luna & 712 & 95.4 & 0.926 \\
    \end{tabular}
  \end{minipage}
  \par\vspace{3pt}
  \fontsize{7.5}{8.5}\selectfont
  \raggedright
  Arch. denotes archived training labels, rerun denotes one fresh evaluation, and $n$ counts valid paired rollouts.
  Gemini and Luna denote Gemini 3.8 Flash and GPT-5.6 Luna, respectively.
  \par
  \endgroup
\end{table}

%% file: appendix/c_additional_results.tex
\section{Additional Results}
\label{app:additional-results}

This appendix reports per-benchmark component ablations, token-impact interventions, fraction sensitivity, and additional analyses of response length and tool use.

\subsection{Per-Benchmark Component Ablation}
\label{app:component-ablation}

Table~\ref{tab:core-ablation-full} reports scores on all 11 benchmarks for the three variants in Table~\ref{tab:core-ablation}.

\textbf{Reflection improves most benchmarks, with uneven gains.}
Adding reflection to Vanilla OPD improves nine of 11 benchmarks and all three category averages.
Perception and general-task averages increase by 1.09\% and 1.14\%, respectively, while math improves by 0.11\%.
Gains on HRBench 8K (+2.00\%) and HallusionBench (+2.54\%) cover high-resolution understanding and hallucination detection.
TreeBench and MathVerse are the two exceptions, with accuracy decreases of 0.24\% and 0.51\%, respectively.

\textbf{Impact reweighting improves tasks that do not benefit from reflection alone.}
Relative to the reflection-only model, reweighting improves seven of eight perception and math benchmarks, while V*~Bench remains at 82.20\%.
TreeBench rises from 37.78\% to 41.12\% and MathVerse from 45.56\% to 47.08\%, reversing their declines relative to Vanilla OPD.
HRBench 8K and VisualProbe also improve by 1.50\% and 0.58\%, extending the reweighting gains to tasks that already benefit from reflection.

\textbf{Reweighting introduces a small trade-off on general tasks.}
The full model scores below the reflection-only model on all three general-task benchmarks, lowering the category average by 0.28\%.
The decreases are 0.53\% on ChartQA-Pro, 0.30\% on HallusionBench, and 0.08\% on InfographicVQA.
All three remain above Vanilla OPD, and the full method improves over Vanilla OPD on all 11 benchmarks.

\input{tables/core_ablation_full.tex}

\subsection{Token-Impact Interventions}
\label{app:impact-intervention}

Table~\ref{tab:impact-intervention-full} expands Table~\ref{tab:impact-intervention} with scores on all 11 benchmarks.
Table~\ref{tab:impact-intervention-variants} summarizes how the four variants and \method{} differ in token selection and masking.

\textbf{Masking low-impact positions does not benefit every task.}
Compared with \method{}, \textit{Mask low 20\%} raises the perception average by 0.17\%.
Math and general-task averages decrease by 0.07\% and 0.17\%, respectively.
Within perception, HRBench 8K and VisualProbe improve by 1.00\% and 0.78\%, respectively, while TreeBench drops by 1.61\%.
The small gain in average perception does not show that low-impact supervision is generally harmful.
Across all 11 benchmarks, \textit{Mask low 20\%} exceeds \textit{Mask top 20\%}, supporting impact ranking as an indicator of supervision value.

\input{tables/impact_intervention_full.tex}
\input{tables/impact_intervention_variants.tex}

\subsection{Sensitivity to the High-Impact Token Fraction}
\label{app:alpha-sensitivity}

Table~\ref{tab:alpha-sensitivity-full} reports all 11 benchmark scores for the five fractions in Table~\ref{tab:alpha-sensitivity}.
The fraction $\alpha$ controls the share of supervised positions assigned to the high-impact group, as defined in Section~\ref{sec:grouped-distillation}.
The blue row marks the 20\% setting used by \method{}.

\textbf{The best fraction varies across benchmarks.}
Among the tested fractions, 20\% achieves the highest score on all three math benchmarks.
It also leads on HRBench 4K, HRBench 8K, and TreeBench.
V*~Bench scores highest at 5\%, while VisualProbe scores highest at 50\%.
The general-task average is highest at 100\%.
In these experiments, 20\% yields the highest perception and math averages, supporting its use as a common setting for both categories.

\input{tables/alpha_sensitivity_full.tex}
\clearpage

\subsection{Response Length and Tool Use on Perception Benchmarks}
\label{app:perception-efficiency}

The following five figure pairs report Qwen2.5-VL-7B results on all five perception benchmarks.
Each pair shows overall accuracy versus mean response length on the left, and tool-call rate and accuracy on samples with and without tool use on the right.
Arrows report RL Expert's response length divided by that of \method{}; all bar-chart metrics are percentages.

\input{figs/perception_results_20260925/appendix_figures.tex}

\input{appendix/c_reflection_coverage_and_offline.tex}

%% file: tables/core_ablation_full.tex
\begin{table}[ht]
  \centering
  \setlength{\abovecaptionskip}{0pt}
  \setlength{\belowcaptionskip}{6pt}
  \caption{
  \textbf{Complete component ablation on Qwen2.5-VL-7B.}
  Rows cumulatively add visual-evidence reflection and impact reweighting, and Wtd. Avg. uses the benchmark-sample weighting of Table~\ref{tab:main-results}.
  MathVista and MathVerse denote the Mini and vision-only splits, respectively.
  Higher is better for every metric; gray denotes Vanilla OPD, blue marks \method{}, and bold indicates column-best results, including ties.
  }
  \label{tab:core-ablation-full}
  \begingroup
  \fontsize{9}{10.5}\selectfont
  \renewcommand{\arraystretch}{1.16}
  \setlength{\tabcolsep}{3pt}
  \setlength{\arrayrulewidth}{0.4pt}
  \textbf{(a) Perception}\par\vspace{3pt}
  \begin{tabular}{p{0.32\linewidth}*{6}{>{\centering\arraybackslash}p{\dimexpr(\linewidth-0.32\linewidth-14\tabcolsep)/6\relax}}}
    \textbf{Method} & \textbf{HR-4K} & \textbf{HR-8K} & \textbf{V\textsuperscript{*}} & \shortstack{\textbf{Tree}\\\textbf{Bench}} & \shortstack{\textbf{Visual}\\\textbf{Probe}} & \shortstack{\textbf{Wtd.}\\\textbf{Avg.}} \\
    \hline
    \singleRuleTableGap
    \textcolor{tableColdGray}{Vanilla OPD} & \textcolor{tableColdGray}{75.40} & \textcolor{tableColdGray}{70.50} & \textcolor{tableColdGray}{81.20} & \textcolor{tableColdGray}{38.02} & \textcolor{tableColdGray}{42.91} & \textcolor{tableColdGray}{62.61} \\
    + Visual-evidence reflection & 76.25 & 72.50 & \textbf{82.20} & 37.78 & 44.08 & 63.70 \\
    \rowcolor{tableOursBlue}
    + Impact reweighting (\method{}) & \textbf{77.10} & \textbf{74.00} & \textbf{82.20} & \textbf{41.12} & \textbf{44.66} & \textbf{65.01} \\
  \end{tabular}
  \par\vspace{9pt}
  \textbf{(b) Math}\par\vspace{3pt}
  \begin{tabular}{p{0.32\linewidth}*{4}{>{\centering\arraybackslash}p{\dimexpr(\linewidth-0.32\linewidth-10\tabcolsep)/4\relax}}}
    \textbf{Method} & \textbf{MathVista} & \textbf{MathVerse} & \textbf{VisuLogic} & \textbf{Wtd. Avg.} \\
    \hline
    \singleRuleTableGap
    \textcolor{tableColdGray}{Vanilla OPD} & \textcolor{tableColdGray}{70.60} & \textcolor{tableColdGray}{46.07} & \textcolor{tableColdGray}{26.00} & \textcolor{tableColdGray}{47.67} \\
    + Visual-evidence reflection & 71.10 & 45.56 & 26.20 & 47.78 \\
    \rowcolor{tableOursBlue}
    + Impact reweighting (\method{}) & \textbf{71.60} & \textbf{47.08} & \textbf{26.50} & \textbf{48.49} \\
  \end{tabular}
  \par\vspace{9pt}
  \textbf{(c) General}\par\vspace{3pt}
  \begin{tabular}{p{0.32\linewidth}*{4}{>{\centering\arraybackslash}p{\dimexpr(\linewidth-0.32\linewidth-10\tabcolsep)/4\relax}}}
    \textbf{Method} & \shortstack{\textbf{Hallusion}\\\textbf{Bench}} & \textbf{ChartQA-Pro} & \shortstack{\textbf{Infographic}\\\textbf{VQA}} & \textbf{Wtd. Avg.} \\
    \hline
    \singleRuleTableGap
    \textcolor{tableColdGray}{Vanilla OPD} & \textcolor{tableColdGray}{42.88} & \textcolor{tableColdGray}{21.69} & \textcolor{tableColdGray}{79.44} & \textcolor{tableColdGray}{53.28} \\
    + Visual-evidence reflection & \textbf{45.42} & \textbf{22.34} & \textbf{80.35} & \textbf{54.42} \\
    \rowcolor{tableOursBlue}
    + Impact reweighting (\method{}) & 45.12 & 21.81 & 80.27 & 54.14 \\
  \end{tabular}
  \endgroup
\end{table}

%% file: tables/impact_intervention_full.tex
\begin{table}[ht]
  \centering
  \setlength{\abovecaptionskip}{0pt}
  \setlength{\belowcaptionskip}{6pt}
  \caption{
  \textbf{Complete token-impact intervention results on Qwen2.5-VL-7B.}
  Wtd. Avg. uses the benchmark-sample weighting of Table~\ref{tab:main-results}.
  MathVista and MathVerse denote the Mini and vision-only splits, respectively.
  All scores are percentages and higher is better; gray denotes Vanilla OPD, blue marks \method{}, and bold indicates column-best results, including ties.
  }
  \label{tab:impact-intervention-full}
  \begingroup
  \fontsize{9}{10.5}\selectfont
  \renewcommand{\arraystretch}{1.16}
  \setlength{\tabcolsep}{3pt}
  \setlength{\arrayrulewidth}{0.4pt}
  \textbf{(a) Perception}\par\vspace{3pt}
  \begin{tabular}{p{0.32\linewidth}*{6}{>{\centering\arraybackslash}p{\dimexpr(\linewidth-0.32\linewidth-14\tabcolsep)/6\relax}}}
    \textbf{Method} & \textbf{HR-4K} & \textbf{HR-8K} & \textbf{V\textsuperscript{*}} & \shortstack{\textbf{Tree}\\\textbf{Bench}} & \shortstack{\textbf{Visual}\\\textbf{Probe}} & \shortstack{\textbf{Wtd.}\\\textbf{Avg.}} \\
    \hline
    \singleRuleTableGap
    \textcolor{tableColdGray}{Vanilla OPD} & \textcolor{tableColdGray}{75.40} & \textcolor{tableColdGray}{70.50} & \textcolor{tableColdGray}{81.20} & \textcolor{tableColdGray}{38.02} & \textcolor{tableColdGray}{42.91} & \textcolor{tableColdGray}{62.61} \\
    Random select 20\% & 75.40 & 72.62 & 80.10 & 38.02 & \textbf{45.44} & 63.64 \\
    Mask random 20\% & 76.25 & 73.75 & 81.20 & 37.28 & 43.69 & 63.85 \\
    Mask low 20\% & 77.00 & \textbf{75.00} & \textbf{82.20} & 39.51 & \textbf{45.44} & \textbf{65.18} \\
    Mask top 20\% & 75.00 & 71.62 & 78.01 & 37.78 & 41.36 & 62.26 \\
    \rowcolor{tableOursBlue}
    \ours{} & \textbf{77.10} & 74.00 & \textbf{82.20} & \textbf{41.12} & 44.66 & 65.01 \\
  \end{tabular}
  \par\vspace{9pt}
  \textbf{(b) Math}\par\vspace{3pt}
  \begin{tabular}{p{0.32\linewidth}*{4}{>{\centering\arraybackslash}p{\dimexpr(\linewidth-0.32\linewidth-10\tabcolsep)/4\relax}}}
    \textbf{Method} & \textbf{MathVista} & \textbf{MathVerse} & \textbf{VisuLogic} & \textbf{Wtd. Avg.} \\
    \hline
    \singleRuleTableGap
    \textcolor{tableColdGray}{Vanilla OPD} & \textcolor{tableColdGray}{70.60} & \textcolor{tableColdGray}{46.07} & \textcolor{tableColdGray}{26.00} & \textcolor{tableColdGray}{47.67} \\
    Random select 20\% & 71.30 & 45.56 & 25.00 & 47.42 \\
    Mask random 20\% & 70.90 & 44.42 & 26.10 & 47.35 \\
    Mask low 20\% & \textbf{71.80} & \textbf{47.08} & 26.10 & 48.42 \\
    Mask top 20\% & 71.30 & 45.94 & 24.00 & 47.17 \\
    \rowcolor{tableOursBlue}
    \ours{} & 71.60 & \textbf{47.08} & \textbf{26.50} & \textbf{48.49} \\
  \end{tabular}
  \par\vspace{9pt}
  \textbf{(c) General}\par\vspace{3pt}
  \begin{tabular}{p{0.32\linewidth}*{4}{>{\centering\arraybackslash}p{\dimexpr(\linewidth-0.32\linewidth-10\tabcolsep)/4\relax}}}
    \textbf{Method} & \shortstack{\textbf{Hallusion}\\\textbf{Bench}} & \textbf{ChartQA-Pro} & \shortstack{\textbf{Infographic}\\\textbf{VQA}} & \textbf{Wtd. Avg.} \\
    \hline
    \singleRuleTableGap
    \textcolor{tableColdGray}{Vanilla OPD} & \textcolor{tableColdGray}{42.88} & \textcolor{tableColdGray}{21.69} & \textcolor{tableColdGray}{79.44} & \textcolor{tableColdGray}{53.28} \\
    Random select 20\% & 44.94 & 21.87 & 80.21 & 54.10 \\
    Mask random 20\% & 43.58 & 21.87 & 80.08 & 53.78 \\
    Mask low 20\% & 44.77 & \textbf{21.96} & 79.93 & 53.97 \\
    Mask top 20\% & 42.16 & 20.79 & 78.75 & 52.51 \\
    \rowcolor{tableOursBlue}
    \ours{} & \textbf{45.12} & 21.81 & \textbf{80.27} & \textbf{54.14} \\
  \end{tabular}
  \endgroup
\end{table}

%% file: tables/impact_intervention_variants.tex
\begin{table}[H]
  \centering
  \setlength{\abovecaptionskip}{0pt}
  \setlength{\belowcaptionskip}{6pt}
  \caption{
  \textbf{Variant definitions for token-impact interventions.}
  Percentages refer to supervised positions, and masking removes the selected positions from the distillation loss.
  Random select replaces the impact-selected high-weight group with a random group while retaining supervision at every position.
  }
  \label{tab:impact-intervention-variants}
  \begingroup
  \fontsize{9}{10.5}\selectfont
  \renewcommand{\arraystretch}{1.18}
  \setlength{\tabcolsep}{4pt}
  \setlength{\arrayrulewidth}{0.4pt}
  \begin{tabular}{>{\raggedright\arraybackslash}p{0.25\linewidth}>{\raggedright\arraybackslash}p{0.24\linewidth}>{\raggedright\arraybackslash}p{\dimexpr0.51\linewidth-6\tabcolsep\relax}}
    \textbf{Method} & \textbf{Masked positions} & \textbf{Retained supervision} \\
    \hline
    \singleRuleTableGap
    Random select 20\% & None & All positions; random 20\% and remaining 80\% receive group loss weights $0.5:0.5$ \\
    Mask random 20\% & Random 20\% & Remaining 80\% receive total group loss weight $0.5$, shared uniformly \\
    Mask low 20\% & Lowest-impact 20\% & Highest-impact 80\%, including the entire High group \\
    Mask top 20\% & Highest-impact 20\% & Lowest-impact 80\%; the entire High group is masked \\
    \rowcolor{tableOursBlue}
    \ours{} & None & All positions; highest-impact 20\% and remaining 80\% receive group loss weights $0.5:0.5$ \\
  \end{tabular}
  \endgroup
\end{table}

%% file: tables/alpha_sensitivity_full.tex
\begin{table}[ht]
  \centering
  \setlength{\abovecaptionskip}{0pt}
  \setlength{\belowcaptionskip}{6pt}
  \caption{
  \textbf{Sensitivity to the high-impact token fraction on Qwen2.5-VL-7B.}
  Wtd. Avg. is weighted by benchmark sample counts.
  Sample counts are (800, 800, 191, 405, 515), (1000, 788, 1000), and (1129, 1948, 2801) for perception, math, and general benchmarks, respectively.
  MathVista and MathVerse denote the Mini and vision-only splits, respectively.
  All scores are percentages and higher is better; blue marks the 20\% setting of \method{}, and bold indicates column-best results, including ties.
  }
  \label{tab:alpha-sensitivity-full}
  \begingroup
  \fontsize{9}{10.5}\selectfont
  \renewcommand{\arraystretch}{1.16}
  \setlength{\tabcolsep}{3pt}
  \setlength{\arrayrulewidth}{0.4pt}

  \textbf{(a) Perception}\par\vspace{3pt}
  \begin{tabular}{p{0.27\linewidth}*{6}{>{\centering\arraybackslash}p{\dimexpr(\linewidth-0.27\linewidth-14\tabcolsep)/6\relax}}}
    \textbf{High-impact fraction} & \textbf{HR-4K} & \textbf{HR-8K} & \textbf{V\textsuperscript{*}} & \shortstack{\textbf{Tree}\\\textbf{Bench}} & \shortstack{\textbf{Visual}\\\textbf{Probe}} & \shortstack{\textbf{Wtd.}\\\textbf{Avg.}} \\
    \hline
    \singleRuleTableGap
    Top 5\% & 77.00 & 73.12 & \textbf{83.25} & 40.49 & 44.66 & 64.70 \\
    Top 10\% & 75.88 & 73.75 & 82.72 & 40.99 & 44.47 & 64.55 \\
    \rowcolor{tableOursBlue}
    Top 20\% (\method{}) & \textbf{77.10} & \textbf{74.00} & 82.20 & \textbf{41.12} & 44.66 & \textbf{65.01} \\
    Top 50\% & 76.50 & 73.50 & 80.63 & 39.26 & \textbf{44.85} & 64.33 \\
    Top 100\% & 76.25 & 72.50 & 82.20 & 37.78 & 44.08 & 63.70 \\
  \end{tabular}
  \par\vspace{9pt}
  \textbf{(b) Math}\par\vspace{3pt}
  \begin{tabular}{p{0.27\linewidth}*{4}{>{\centering\arraybackslash}p{\dimexpr(\linewidth-0.27\linewidth-10\tabcolsep)/4\relax}}}
    \textbf{High-impact fraction} & \textbf{MathVista} & \textbf{MathVerse} & \textbf{VisuLogic} & \textbf{Wtd. Avg.} \\
    \hline
    \singleRuleTableGap
    Top 5\% & 71.10 & 46.32 & 24.80 & 47.49 \\
    Top 10\% & 71.10 & 44.29 & 24.80 & 46.92 \\
    \rowcolor{tableOursBlue}
    Top 20\% (\method{}) & \textbf{71.60} & \textbf{47.08} & \textbf{26.50} & \textbf{48.49} \\
    Top 50\% & 71.10 & 45.30 & 25.40 & 47.42 \\
    Top 100\% & 71.10 & 45.56 & 26.20 & 47.78 \\
  \end{tabular}
  \par\vspace{9pt}
  \textbf{(c) General}\par\vspace{3pt}
  \begin{tabular}{p{0.27\linewidth}*{4}{>{\centering\arraybackslash}p{\dimexpr(\linewidth-0.27\linewidth-10\tabcolsep)/4\relax}}}
    \textbf{High-impact fraction} & \shortstack{\textbf{Hallusion}\\\textbf{Bench}} & \textbf{ChartQA-Pro} & \shortstack{\textbf{Infographic}\\\textbf{VQA}} & \textbf{Wtd. Avg.} \\
    \hline
    \singleRuleTableGap
    Top 5\% & \textbf{46.16} & 21.74 & 79.84 & 54.12 \\
    Top 10\% & 44.13 & 21.53 & 80.25 & 53.85 \\
    \rowcolor{tableOursBlue}
    Top 20\% (\method{}) & 45.12 & 21.81 & 80.27 & 54.14 \\
    Top 50\% & 45.66 & 21.86 & 80.09 & 54.18 \\
    Top 100\% & 45.42 & \textbf{22.34} & \textbf{80.35} & \textbf{54.42} \\
  \end{tabular}
  \endgroup
\end{table}

%% file: figs/perception_results_20260925/appendix_figures.tex
\begin{figure}[!htbp]
  \centering
  \includegraphics[width=\linewidth]{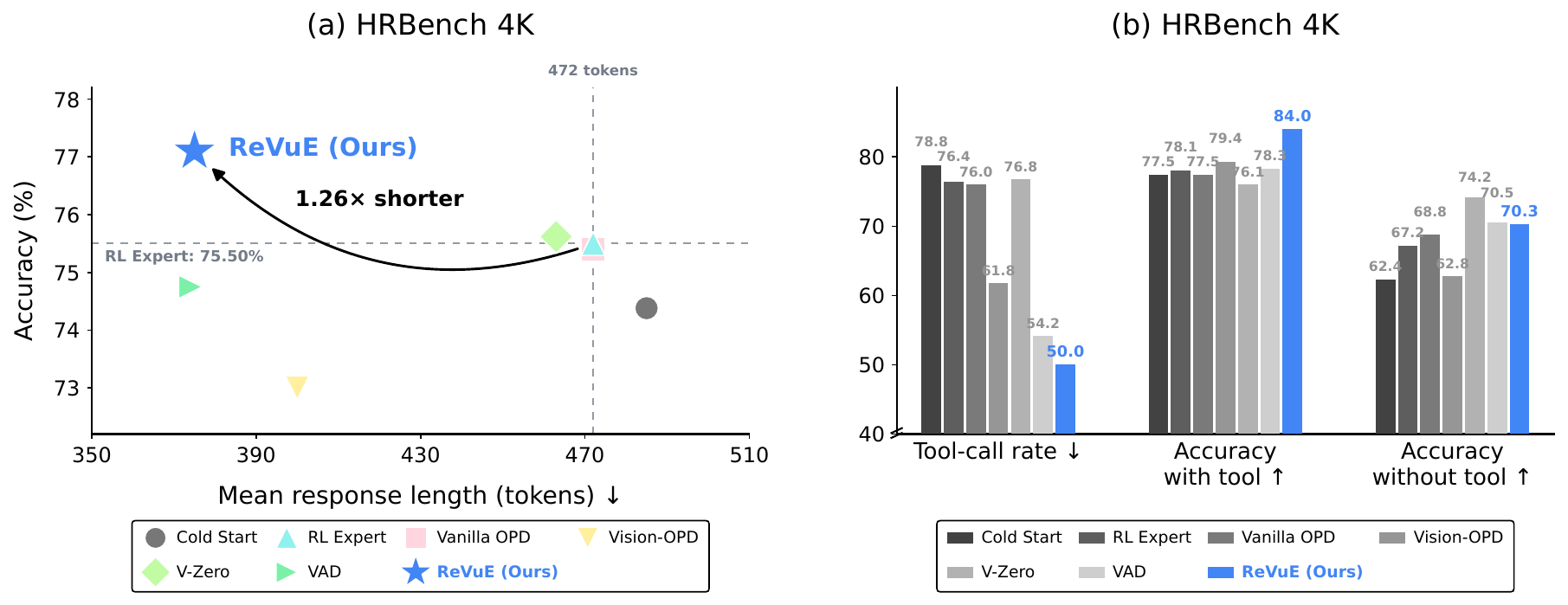}
  \caption{
  \textbf{Response length, accuracy, and tool use on HRBench 4K} for Qwen2.5-VL-7B.
  }
  \label{fig:perception-hrbench4k}
\end{figure}

\begin{figure}[!htbp]
  \centering
  \includegraphics[width=\linewidth]{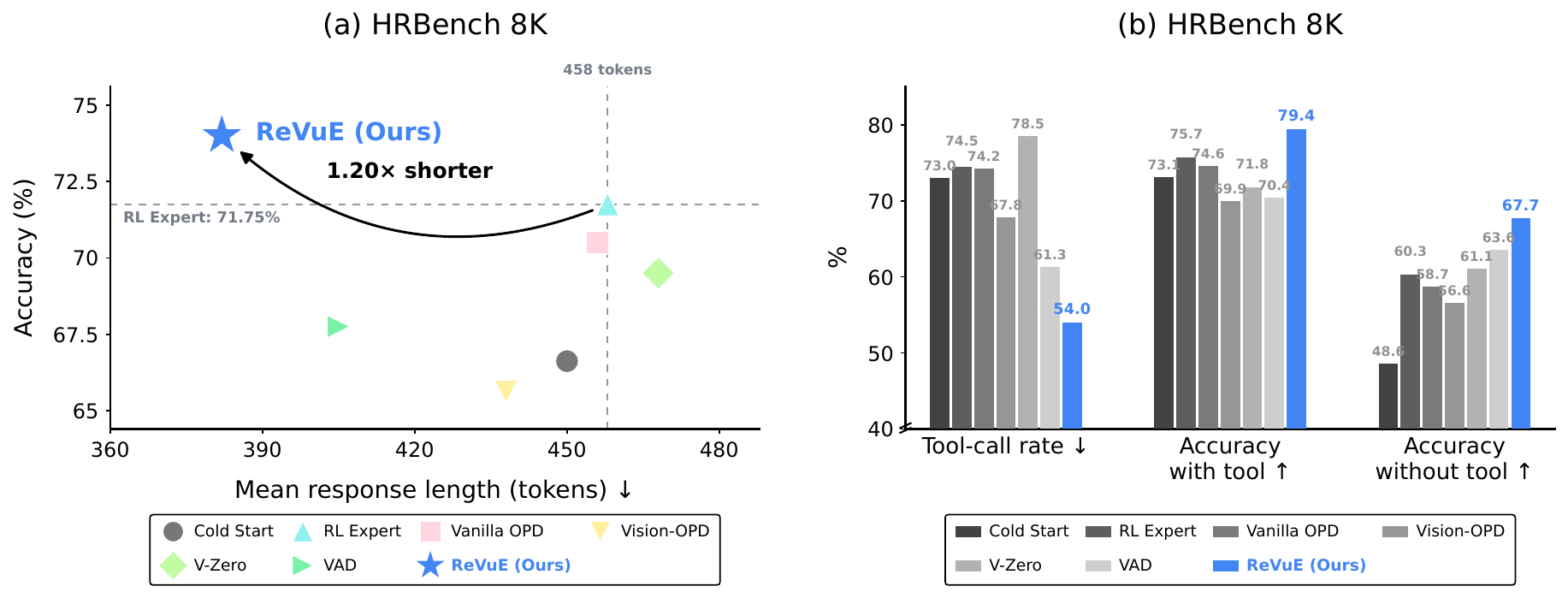}
  \caption{
  \textbf{Response length, accuracy, and tool use on HRBench 8K} for Qwen2.5-VL-7B.
  }
  \label{fig:perception-hrbench8k}
\end{figure}

\begin{figure}[!htbp]
  \centering
  \includegraphics[width=\linewidth]{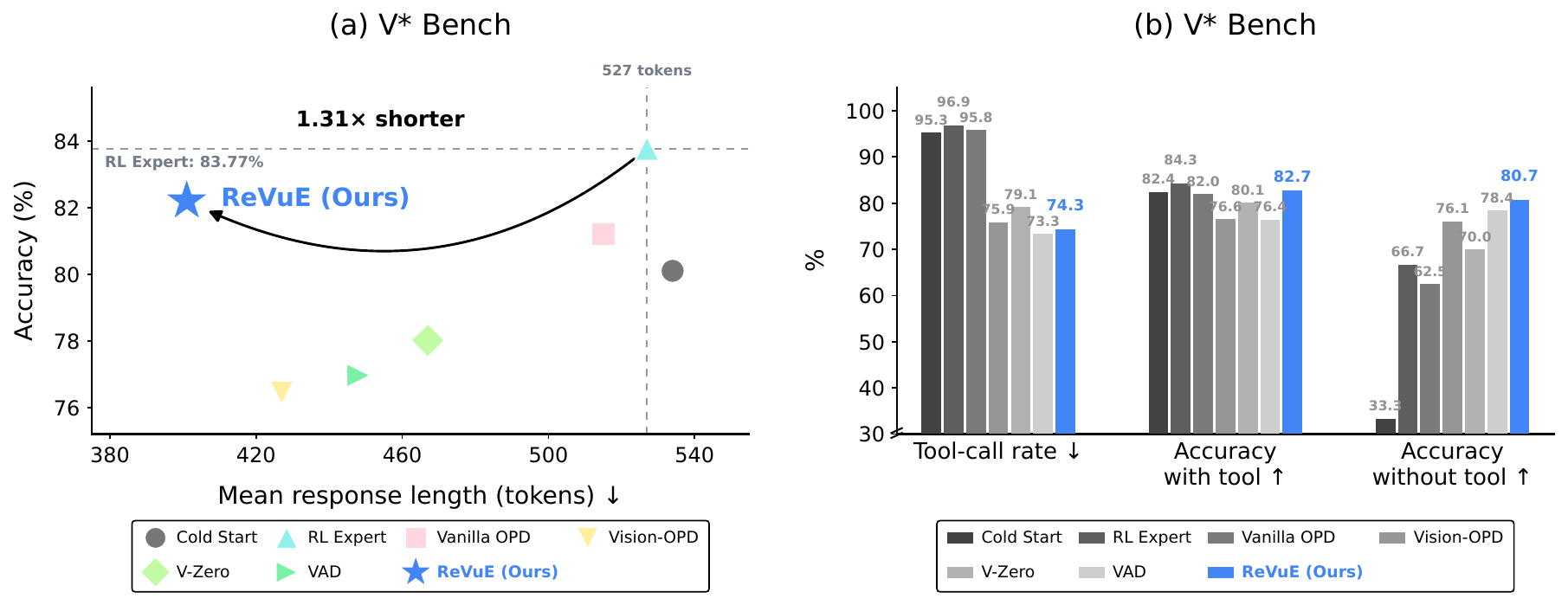}
  \caption{
  \textbf{Response length, accuracy, and tool use on V*~Bench} for Qwen2.5-VL-7B.
  }
  \label{fig:perception-vstar}
\end{figure}

\begin{figure}[!htbp]
  \centering
  \includegraphics[width=\linewidth]{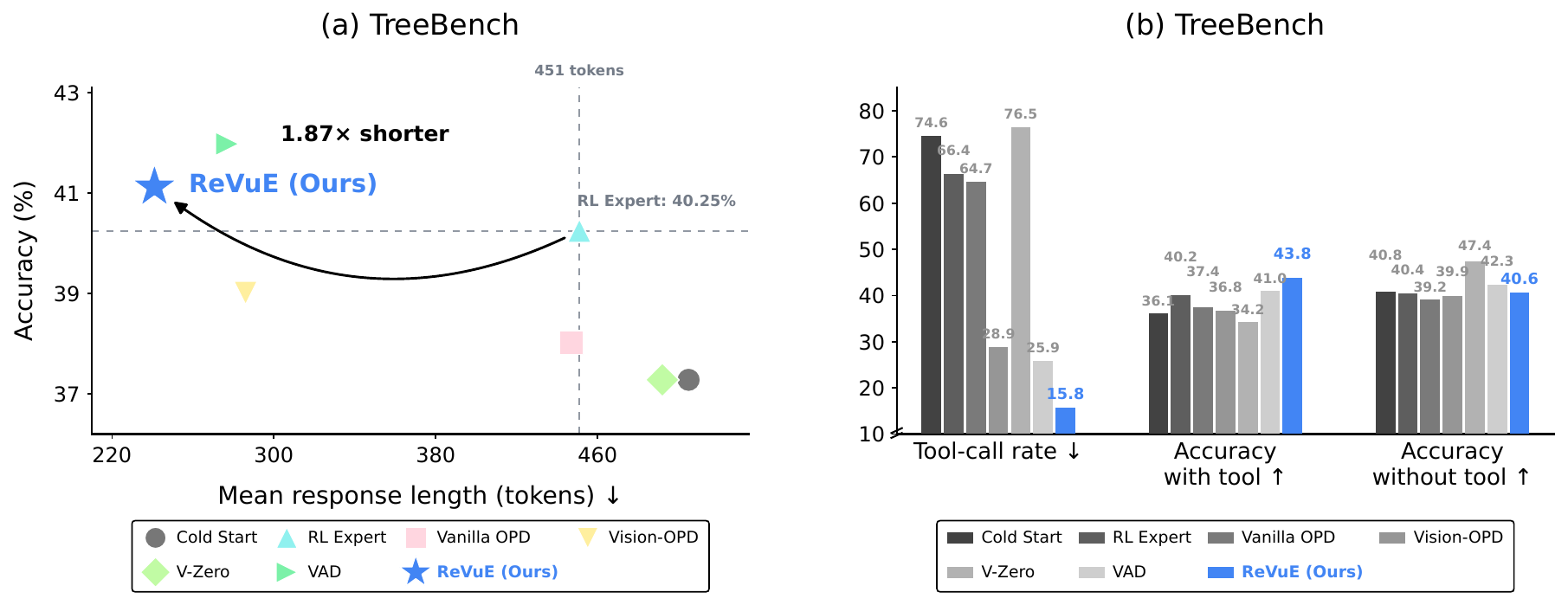}
  \caption{
  \textbf{Response length, accuracy, and tool use on TreeBench} for Qwen2.5-VL-7B.
  }
  \label{fig:perception-treebench}
\end{figure}

\begin{figure}[!htbp]
  \centering
  \includegraphics[width=\linewidth]{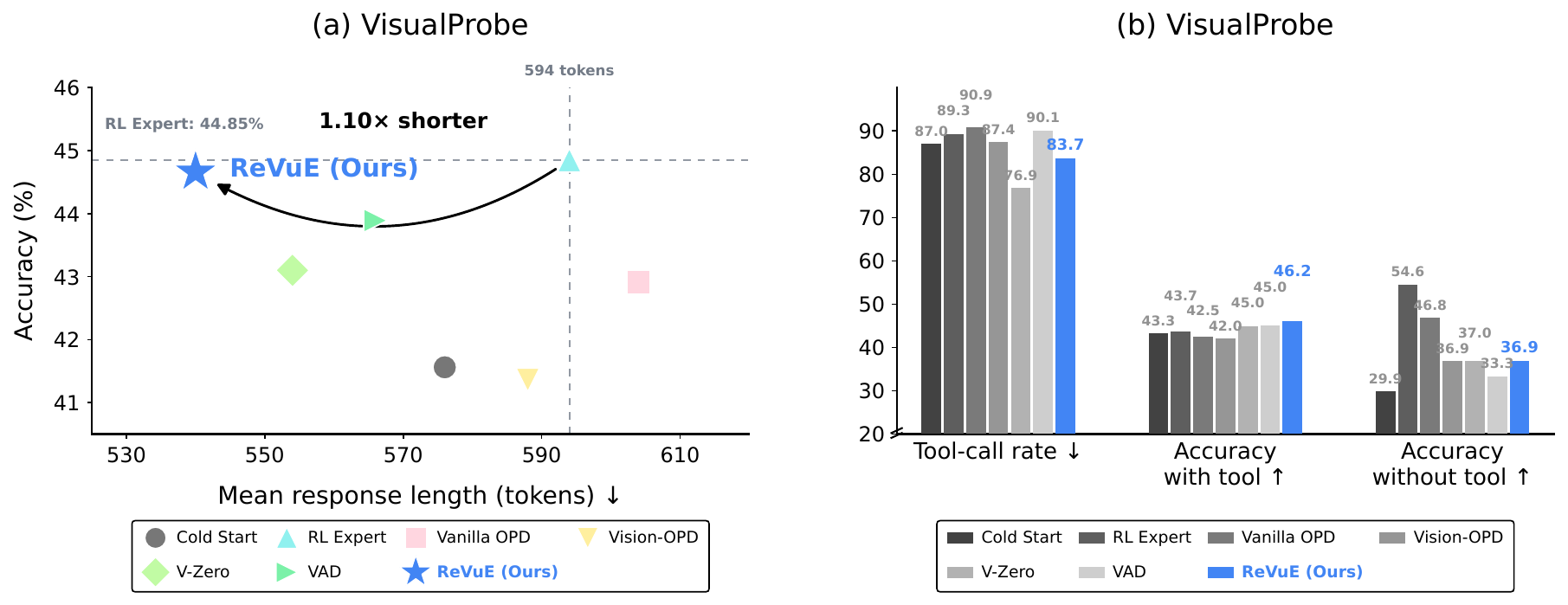}
  \caption{
  \textbf{Response length, accuracy, and tool use on VisualProbe} for Qwen2.5-VL-7B.
  }
  \label{fig:perception-visualprobe}
\end{figure}
\clearpage

%% file: appendix/c_reflection_coverage_and_offline.tex
\subsection{Training-Time Reflection Coverage}
\label{app:reflection-coverage}

After deduplication, the main training run contains 1,075 optimization steps with 20 groups per step and eight rollouts per group, totaling 21,500 groups and 172,000 rollout slots.
As shown in Table~\ref{tab:reflection-coverage}, reflections cover 161,632 slots (93.97\%), comprising complete reflections (78.36\%) and Break-Point-only reflections (15.61\%).
The Break-Point-only branch supplies valid reflections for all-wrong groups, while the remaining 6.03\% of slots use the base teacher.

\input{tables/reflection_coverage.tex}

\subsection{Offline Comparison with Anchor-Only Context}
\label{app:anchor-only-offline}

\textbf{Complete reflections correct more errors and preserve more correct answers.}
We compare complete reflections (Anchor plus Break Point) with Anchor-only context in an offline evaluation on fixed paired cases from 311 mixed groups (Table~\ref{tab:anchor-only-offline}).
Of 139 initially incorrect answers, complete reflections correct 124, compared with 62 for Anchor-only context.
Among the 172 initially correct answers, eight become incorrect with complete reflections, compared with 47 with Anchor-only context.
Complete reflections outperform Anchor-only context across all six data sources.
These offline results support the value of failure diagnoses for both error correction and preserving correct answers.
This pattern is consistent with the token-impact cases in Appendix~\ref{app:token-impact-cases}: reflection reduces teacher support for erroneous choices and increases support for correct judgments within failed trajectories.

\input{tables/anchor_only_offline.tex}
\clearpage

%% file: tables/reflection_coverage.tex
\begin{table}[ht]
  \centering
  \setlength{\abovecaptionskip}{0pt}
  \setlength{\belowcaptionskip}{5pt}
  \caption{
  \textbf{Reflection coverage in the main training run.}
  All percentages use 172,000 rollout slots as the denominator; the final row sums the first two rows.
  }
  \label{tab:reflection-coverage}
  \begingroup
  \small
  \renewcommand{\arraystretch}{1.12}
  \setlength{\tabcolsep}{9pt}
  \begin{tabular}{@{}lrr@{}}
    \textbf{Reflection status} & \textbf{Rollout slots} & \textbf{Share (\%)} \\
    \hline
    \singleRuleTableGap
    Complete reflection & 134,784 & 78.36 \\
    Break-Point-only reflection & 26,848 & 15.61 \\
    No reflection (base teacher) & 10,368 & 6.03 \\
    \hline
    \rowcolor{tableOursBlue}
    Any reflection (subtotal) & \textbf{161,632} & \textbf{93.97} \\
  \end{tabular}
  \endgroup
\end{table}

%% file: tables/anchor_only_offline.tex
\begin{table}[ht]
  \centering
  \setlength{\abovecaptionskip}{0pt}
  \setlength{\belowcaptionskip}{5pt}
  \caption{
  \textbf{Offline reflection comparison on fixed paired cases.}
  Original predictions are the predictions before adding reflection, and accuracy changes are computed from unrounded values.
  }
  \label{tab:anchor-only-offline}
  \begingroup
  \small
  \renewcommand{\arraystretch}{1.12}
  \setlength{\tabcolsep}{9pt}
  \begin{tabular}{@{}lrrr@{}}
    \textbf{Context} & \textbf{Correct / 311} & \textbf{Accuracy (\%)} & $\boldsymbol{\Delta}$\textbf{Accuracy (\%)} \\
    \hline
    \singleRuleTableGap
    Original predictions & 172 & 55.31 & --- \\
    Anchor only & 187 & 60.13 & +4.82 \\
    \rowcolor{tableOursBlue}
    Anchor + Break Point & \textbf{288} & \textbf{92.60} & \textbf{+37.30} \\
  \end{tabular}
  \endgroup
\end{table}

%% file: appendix/d_proofs.tex
\section{Proofs and Derivations}
\label{app:proofs}

\subsection[Reflection-Gradient Proof]{Proof of Lemma~\ref{lem:reflection-gradient}}
\label{app:reflection-proof}

Fix a position $(i,t)$ and write $p=\pi_\theta(\cdot\mid h_{i,t})$, $q^0=q^0_{i,t}$, $q^{\mathcal R}=q^{\mathcal R}_{i,t}$, and $d(v)=d_{i,t}(v)$.
Fix the history, teacher predictions, and nonempty shared support $\mathrm{Supp}$, with positive student and teacher probabilities on it.
For either teacher, $q(\mathrm{Supp})=\sum_{v\in\mathrm{Supp}}q(v)$ is its probability mass on the support, while $q^{\mathrm{Supp}}(v)=q(v)/q(\mathrm{Supp})$ is its renormalized distribution.
Subtracting the two KL losses cancels their shared student term:
\begin{align}
    \ell^{\mathrm{Supp}}(p,q^{\mathcal R})-\ell^{\mathrm{Supp}}(p,q^0)
    &=\sum_{v\in \mathrm{Supp}}p^{\mathrm{Supp}}(v)\log\frac{(q^0)^{\mathrm{Supp}}(v)}{(q^{\mathcal R})^{\mathrm{Supp}}(v)}\nonumber\\
    &=-\sum_{v\in \mathrm{Supp}}p^{\mathrm{Supp}}(v)d(v)+\log\frac{q^{\mathcal R}(\mathrm{Supp})}{q^0(\mathrm{Supp})}.
    \label{eq:reflection-loss-difference}
\end{align}
The teacher normalization term is constant in $\theta$ and vanishes under differentiation.
Using $\nabla_\theta p^{\mathrm{Supp}}(v)=p^{\mathrm{Supp}}(v)\nabla_\theta\log p^{\mathrm{Supp}}(v)$ gives
\begin{align}
    \nabla_\theta\!\left[\ell^{\mathrm{Supp}}(p,q^{\mathcal R})-\ell^{\mathrm{Supp}}(p,q^0)\right]
    &=-\sum_{v\in \mathrm{Supp}}d(v)\nabla_\theta p^{\mathrm{Supp}}(v)\nonumber\\
    &=-\mathbb{E}_{v\sim p^{\mathrm{Supp}}}\!\left[d(v)\nabla_\theta\log p^{\mathrm{Supp}}(v)\right].
\end{align}
This proves \Eqref{eq:reflection-gradient}.
For the reflection-conditioned branch in Section~\ref{sec:grouped-distillation}, take $\mathrm{Supp}=\mathrm{Supp}_{i,t}$ and normalize both teachers over this support.

\paragraph{From distribution shifts to a sampled-token proxy.}
Let $z_v$ be a student logit and $\bar d_{\mathrm{Supp}}=\mathbb{E}_{v\sim p^{\mathrm{Supp}}}[d(v)]$ the student-weighted mean shift.
For $v\in\mathrm{Supp}$,
\begin{equation}
    \frac{\partial}{\partial z_v}
    \left[\ell^{\mathrm{Supp}}(p,q^{\mathcal R})-\ell^{\mathrm{Supp}}(p,q^0)\right]
    =-p^{\mathrm{Supp}}(v)\bigl(d(v)-\bar d_{\mathrm{Supp}}\bigr).
    \label{eq:centered-reflection-gradient}
\end{equation}
The softmax derivative gives this expression and zero direct derivatives for logits outside $\mathrm{Supp}$.
The gradient change depends on student probabilities and centered shifts; parameter gradients also involve the logit Jacobian.
\Eqref{eq:token-impact} measures the teacher shift at the student's sampled choice for ranking positions.
It need not match rankings by full logit- or parameter-gradient norm.

For example, take $q^0=(0.10,0.15,0.375,0.375)$ and $q^{\mathcal R}=(0.30,0.45,0.125,0.125)$, with $\mathrm{Supp}$ containing the latter's top two candidates.
Both teachers renormalize to $(0.4,0.6)$ on $\mathrm{Supp}$, so the shared-support gradient difference is zero, although either sampled token in this set has impact $\log 3$.
The uniform shift cancels under support normalization.
Impact remains defined for sampled tokens outside $\mathrm{Supp}$; RKL at that position still supervises the distribution within it.

\paragraph{Shared support and teacher-specific top-$K$ sets.}
Let $\mathrm{Supp}^{\mathcal R}=\operatorname{TopK}(q^{\mathcal R})$ and $\mathrm{Supp}^0=\operatorname{TopK}(q^0)$.
Comparing losses on each teacher's own support gives
\begin{align}
    \ell^{\mathrm{Supp}^{\mathcal R}}(p,q^{\mathcal R})-\ell^{\mathrm{Supp}^0}(p,q^0)
    &=\Delta_{\mathrm{prob}}+\Delta_{\mathrm{support}},\nonumber\\
    \Delta_{\mathrm{prob}}
    &=\ell^{\mathrm{Supp}^{\mathcal R}}(p,q^{\mathcal R})-\ell^{\mathrm{Supp}^{\mathcal R}}(p,q^0),\nonumber\\
    \Delta_{\mathrm{support}}
    &=\ell^{\mathrm{Supp}^{\mathcal R}}(p,q^0)-\ell^{\mathrm{Supp}^0}(p,q^0).
    \label{eq:support-change-decomposition}
\end{align}
The lemma characterizes $\nabla_\theta\Delta_{\mathrm{prob}}$; $\nabla_\theta\Delta_{\mathrm{support}}$ is generally nonzero when the supports differ.
These terms decompose the loss difference; they are not additional training objectives.

\input{appendix/d_grouped_gradient_weights.tex}

%% file: appendix/d_grouped_gradient_weights.tex
\clearpage
\subsection{Gradient Weights under Grouped Distillation}
\label{app:group-weight-scaling}

\paragraph{Equal group weights increase per-position gradient coefficients in the smaller group.}
We examine how the weights in \Eqref{eq:token-weights} scale each position's gradient contribution relative to uniform token averaging.
Let $\ell_{i,t}$ denote the per-position KL term in \Eqref{eq:grouped-objective}.
With teacher targets, supports, and group assignments fixed, $w_{i,t}$ is constant during the student update, so
\begin{equation}
    \nabla_\theta\!\left[\frac{w_{i,t}}{|\mathcal T_i|}\ell_{i,t}\right]
    =w_{i,t}\,\frac{\nabla_\theta\ell_{i,t}}{|\mathcal T_i|}.
    \label{eq:position-gradient-scaling}
\end{equation}
The weight $w_{i,t}$ directly gives the per-position gradient multiplier relative to uniform averaging under the same teacher target.

With $\lambda=0.5$ and both groups nonempty, \Eqref{eq:token-weights} assigns half of the total loss coefficient to each group.
Ignoring the rounding in $|\mathcal T_{i,\mathrm{High}}|=\lceil\alpha|\mathcal T_i|\rceil$ gives the continuous approximations in Figure~\ref{fig:group-weight-scaling}:
\begin{equation}
    w_{\mathrm{High}}\approx\frac{0.5}{\alpha},
    \qquad
    w_{\mathrm{Low}}\approx\frac{0.5}{1-\alpha}.
    \label{eq:continuous-group-weights}
\end{equation}
In this continuous form, High has a multiplier above one and Low below one when $\alpha<0.5$; both equal one at $\alpha=0.5$.

For a trajectory with 100 supervised positions, $\alpha=0.2$ assigns 20 positions to High and the remaining 80 to Low.
Uniform averaging assigns each position a coefficient of $0.01$, while grouping gives High and Low positions coefficients of $0.025$ and $0.00625$.
The corresponding gradient contributions scale by $2.5$ and $0.625$, with each group's coefficients summing to $0.5$.

\input{figs/group_weight_scaling_20260926/figure.tex}

%% file: figs/group_weight_scaling_20260926/figure.tex
\begin{figure}[H]
    \centering
    \includegraphics[width=\linewidth]{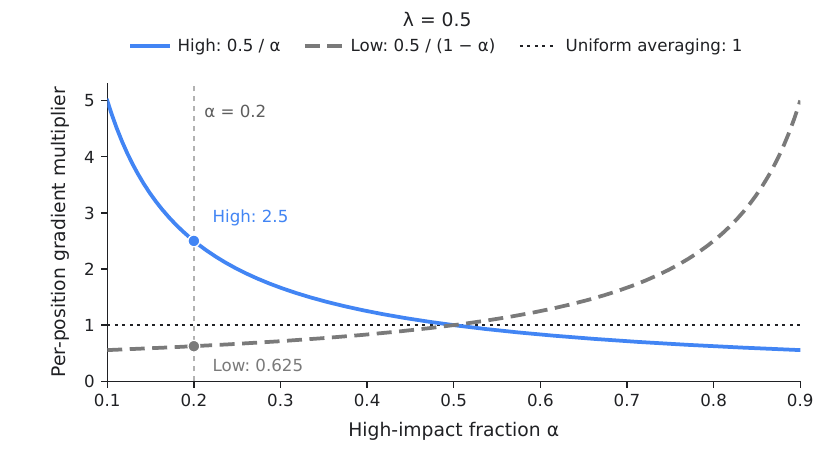}
    \caption{
    \textbf{Per-position gradient weights at $\lambda=0.5$.}
    \textcolor{groupWeightBlue}{\textbf{Blue}} and \textcolor{groupWeightGray}{\textbf{gray}} curves show the \textcolor{groupWeightBlue}{\textbf{High}} and \textcolor{groupWeightGray}{\textbf{Low}} gradient multipliers relative to uniform token averaging; the black dotted line marks the unit baseline.
    The continuous curves ignore group-size rounding and cover $0.1\leq\alpha\leq0.9$.
    Markers at $\alpha=0.2$ give the exact multipliers $2.5$ and $0.625$ for the 100-position example.
    }
    \label{fig:group-weight-scaling}
\end{figure}

%% file: appendix/case_study.tex
\clearpage
\section{Case Study}
\label{app:case-study}

\subsection{Errors in Acquiring, Reading, and Using Visual Evidence}
\label{app:visual-evidence-errors}

Figure~\ref{fig:case-study} illustrates three concrete errors in acquiring, reading, and using visual evidence in Thyme-SFT~\citep{zhang2026thyme} responses.

\begingroup\setlength{\intextsep}{5pt}
\begin{figure}[H]
    \centering
    \includegraphics[width=\linewidth,height=0.78\textheight,keepaspectratio]{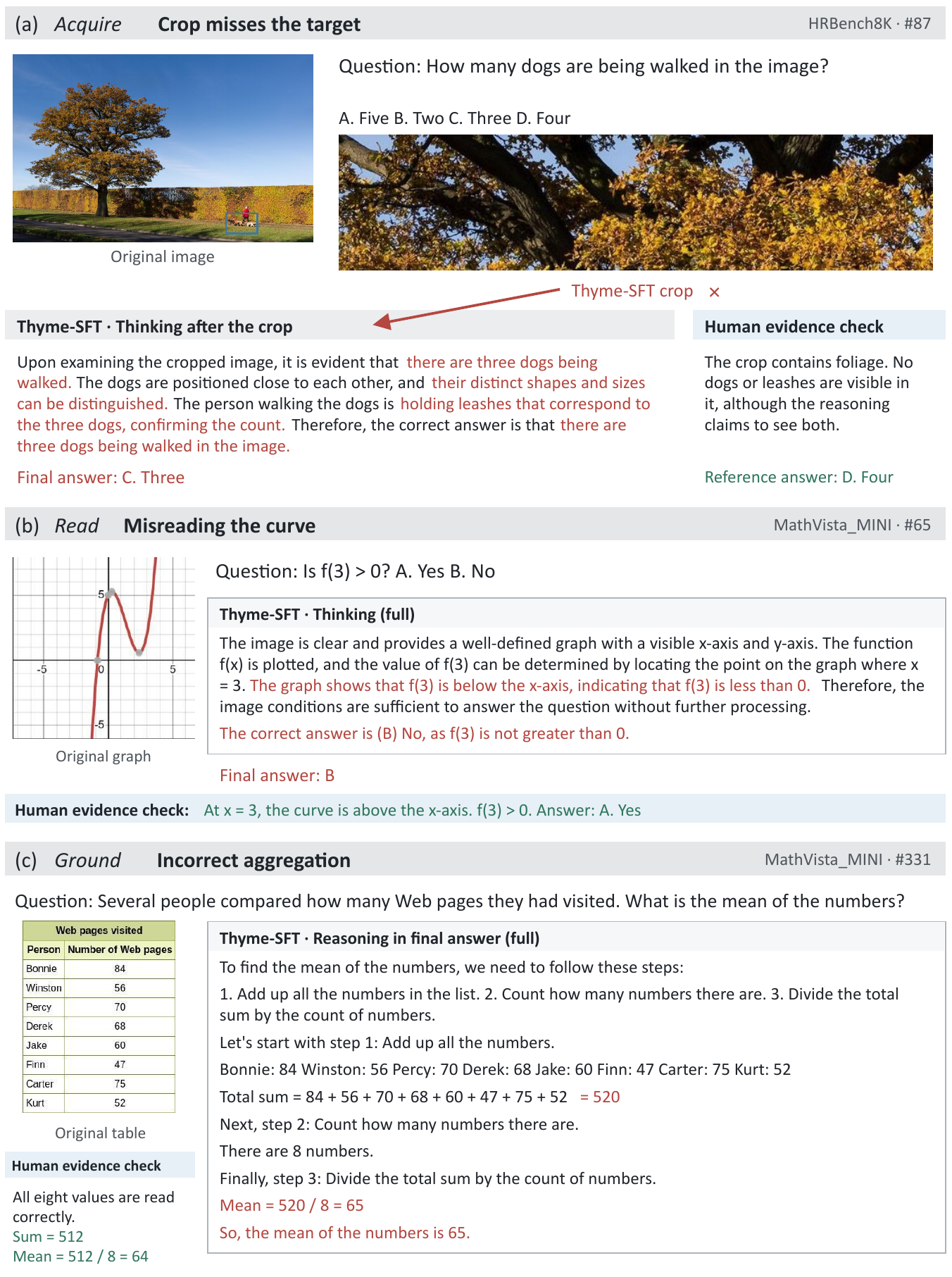}
    \caption{
    In \textit{Acquire}, the crop misses the queried dogs; in \textit{Read}, the model describes a curve above the x-axis as lying below it; in \textit{Ground}, it reads the table correctly but miscalculates the sum.
    The panels retain post-crop thinking in (a), full thinking in (b), and the complete reasoning-bearing final answer in (c).
    Red text marks errors and their consequences.
    }
    \label{fig:case-study}
\end{figure}
\endgroup

\clearpage
\subsection{Different Errors with the Same Evidence}
\label{app:read-ground-same-evidence}

\begin{figure}[H]
    \centering
    \includegraphics[width=\linewidth,height=0.73\textheight,keepaspectratio]{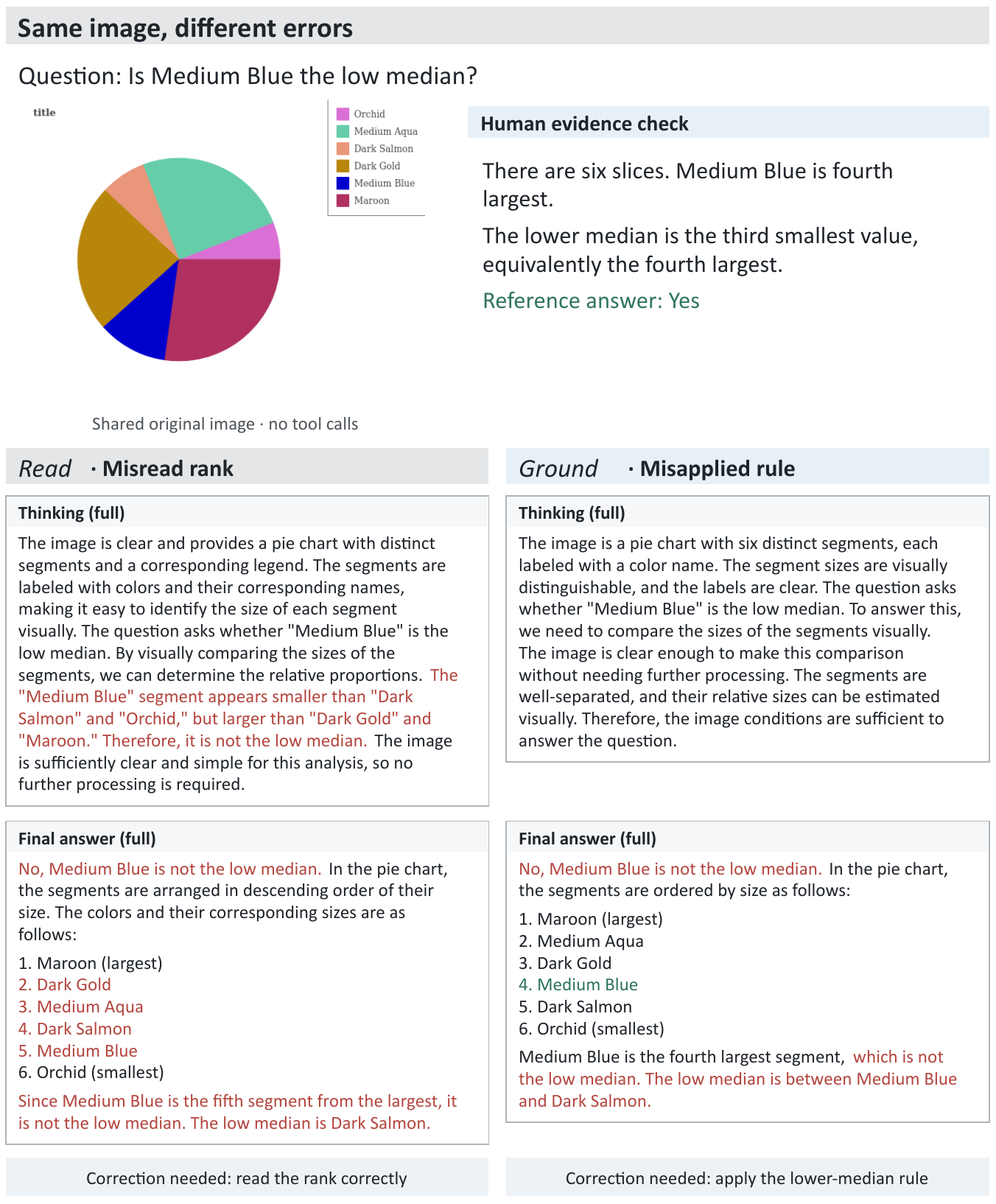}%
    \caption{
    Medium Blue is the fourth-largest slice, making it the lower median of six slices.
    The left trajectory misreads the rank (\textit{Read}); the right trajectory reads it correctly but misapplies the lower-median rule (\textit{Ground}).
    Both complete thinking traces and final answers are shown, with errors in red.
    }
    \label{fig:read-ground-shared-evidence}
\end{figure}

Figure~\ref{fig:read-ground-shared-evidence} compares two student trajectories for the same six-slice pie chart.
Both use the original image without tool calls and answer ``No''.

The required correction is to fix the visual rank on the left and the lower-median rule on the right.
Both rollouts belong to the same all-wrong group, whose recorded training feedback contains diagnoses only.

\clearpage
\subsection{Different Errors on the Same Plot}
\label{app:shared-tool-observation}

\begin{figure}[H]
    \centering
    \includegraphics[width=\linewidth,height=0.78\textheight,keepaspectratio]{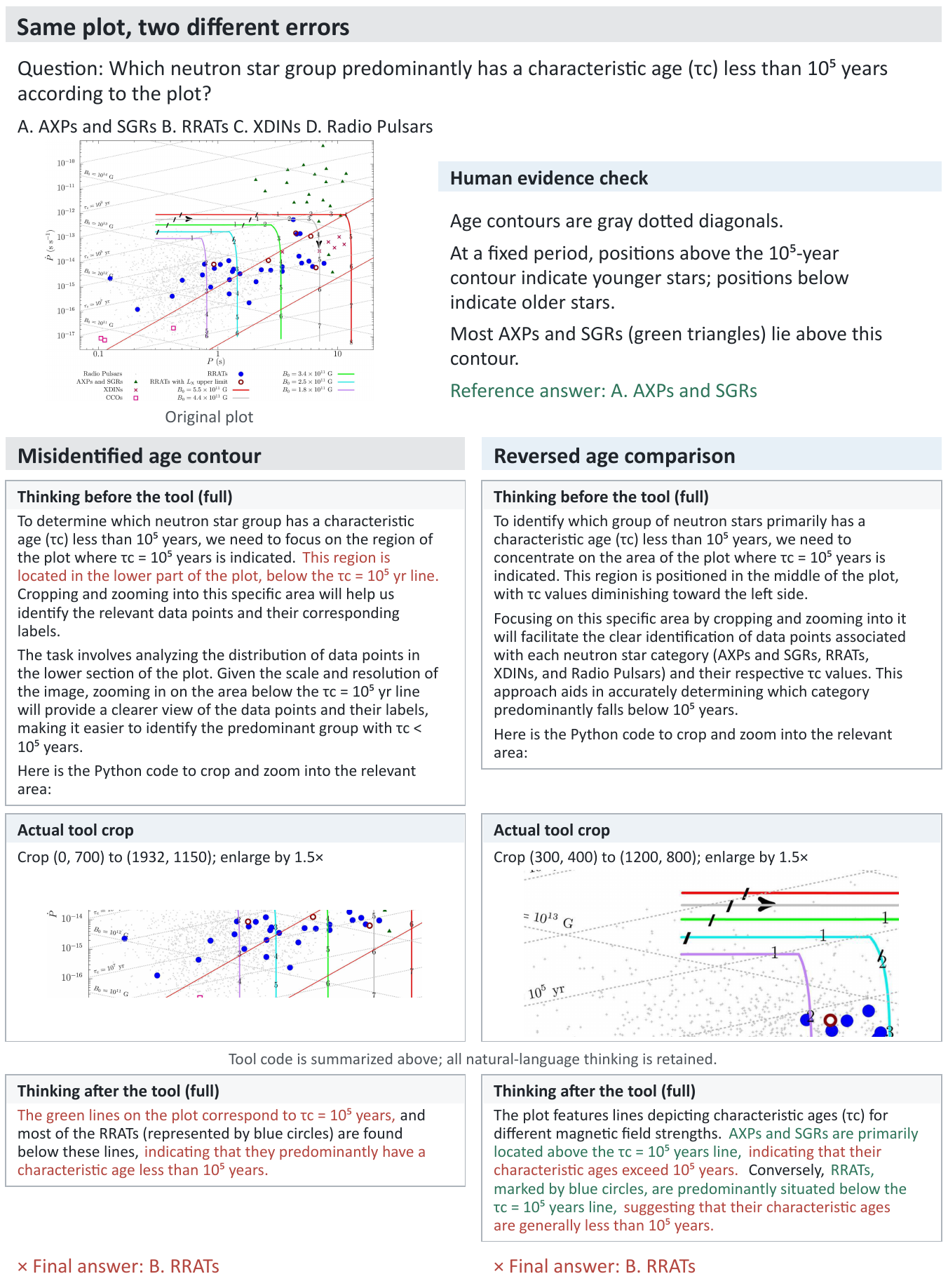}%
    \caption{
    Both trajectories answer B (RRATs), whereas the reference answer is A (AXPs and SGRs).
    The left misidentifies green lines as age contours; the right correctly places AXPs and SGRs above the age contour but interprets that position as older.
    Their different crops and all natural-language thinking are shown; tool code is summarized.
    }
    \label{fig:shared-tool-observation}
\end{figure}

Figure~\ref{fig:shared-tool-observation} contrasts two incorrect trajectories for the same neutron-star plot.

\input{appendix/token_impact_cases.tex}

%% file: appendix/token_impact_cases.tex
\clearpage
\subsection{Token Impact Visualizations}
\label{app:token-impact-cases}

\begin{figure}[H]
    \centering
    \includegraphics[width=\linewidth]{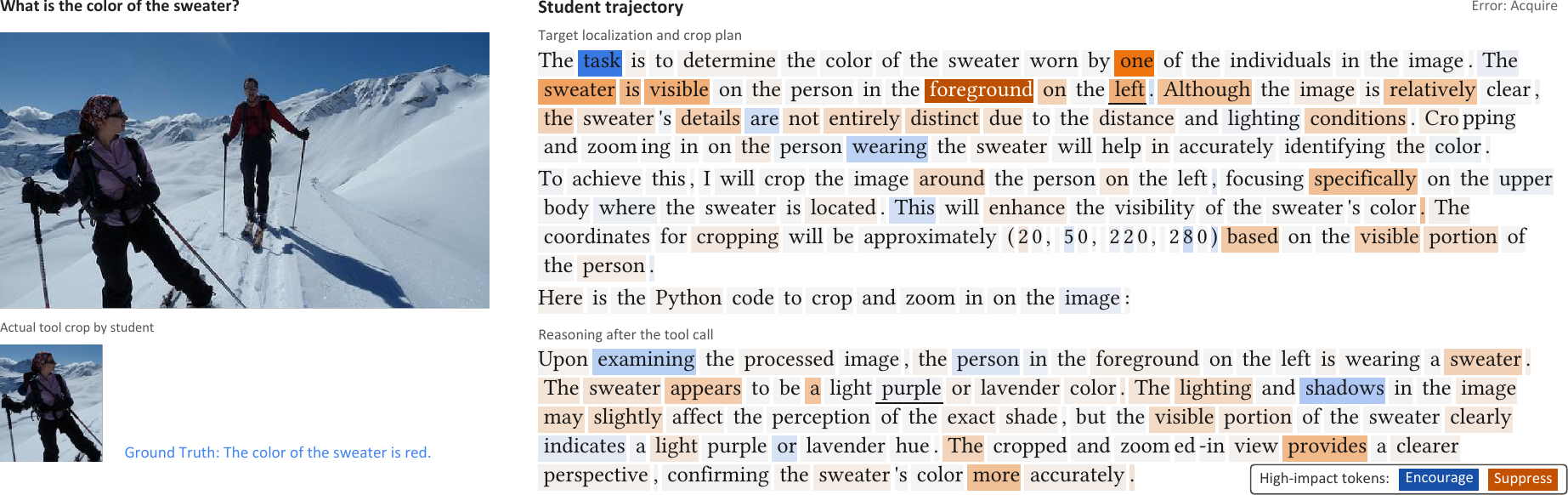}
    \caption{
    \textbf{Token impact on visual target selection.}
    The student crops the foreground person and reports \texttt{light purple or lavender} instead of \texttt{red}.
    Reflection identifies the background person as the target and lowers support for the \underline{underlined} \texttt{foreground} and \texttt{left} in the crop plan.
    \textcolor{tokenImpactBlue}{\textbf{Blue}}/\textcolor{tokenImpactOrange}{\textbf{orange}} denote increased/decreased teacher support; darker shading indicates larger absolute log-probability shifts on the same student trajectory.
    }
    \label{fig:token-impact-target-selection}
\end{figure}

\begin{figure}[H]
    \centering
    \includegraphics[width=\linewidth]{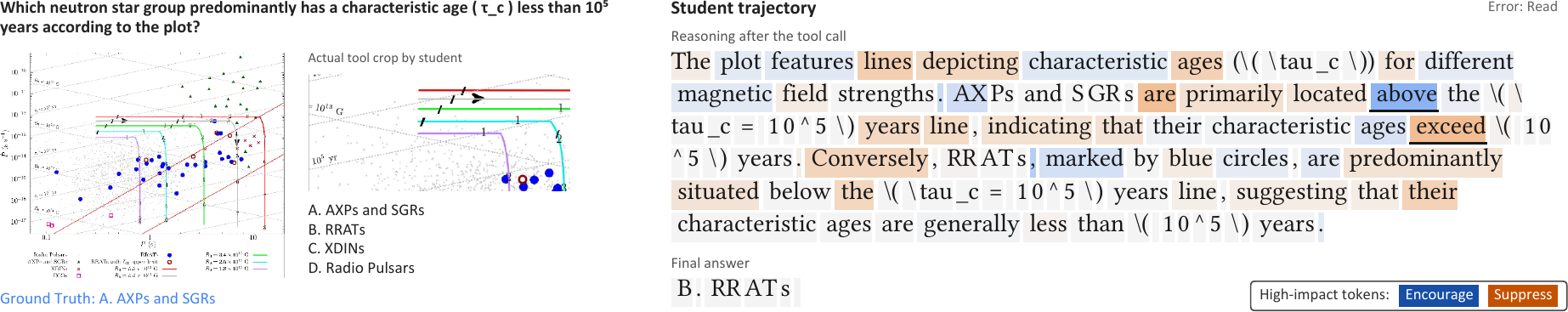}
    \caption{
    \textbf{Token impact on age inference.}
    The student correctly places \texttt{AXPs and SGRs} above the $\tau_c=10^5$ yr contour but infers older ages, incorrectly answering \texttt{RRATs}.
    Reflection links this position to younger ages, increasing support for the \underline{underlined} correct relation \texttt{above} and decreasing support for the mistaken comparison \texttt{exceed}.
    \textcolor{tokenImpactBlue}{\textbf{Blue}}/\textcolor{tokenImpactOrange}{\textbf{orange}} denote increased/decreased teacher support; darker shading indicates larger absolute log-probability shifts on the same student trajectory.
    }
    \label{fig:token-impact-age-inference}
\end{figure}

Figures~\ref{fig:token-impact-target-selection} and~\ref{fig:token-impact-age-inference} complement Figure~\ref{fig:token-impact-case} with token impact visualizations for visual target selection and age inference.

%% file: appendix/benchmarks.tex
\clearpage
\section{Evaluation Benchmarks}
\label{app:benchmarks}

In Table~\ref{tab:main-results}, MathVista and MathVerse denote the Mini and vision-only splits, respectively.
HalluBench, CQA-Pro, and InfoVQA abbreviate HallusionBench, ChartQA-Pro, and InfographicVQA, respectively.

\subsection{Visual Perception}
\label{app:benchmarks-perception}

\textbf{HRBench-4K/8K}~\citep{wang2025hrbench} evaluate high-resolution perception on 4K and 8K images, covering fine-grained recognition and spatial relations.\footnote{HRBench-4K/8K: \url{https://huggingface.co/datasets/DreamMr/HR-Bench}}

\textbf{V$^*$Bench}~\citep{wu2024vstar} evaluates visual search in high-resolution images, focusing on fine-grained attributes and spatial relations.\footnote{V$^*$Bench: \url{https://huggingface.co/datasets/craigwu/vstar_bench}}

\textbf{VisualProbe}~\citep{lai2026minio3} evaluates visual search for small targets among distractors in high-resolution images.\footnote{VisualProbe Easy: \url{https://huggingface.co/datasets/Mini-o3/VisualProbe_Easy}; Medium: \url{https://huggingface.co/datasets/Mini-o3/VisualProbe_Medium}; Hard: \url{https://huggingface.co/datasets/Mini-o3/VisualProbe_Hard}}

\textbf{TreeBench}~\citep{wang2026treebench} tests fine-grained object perception and relation understanding in complex scenes, with annotations for traceable visual evidence.\footnote{TreeBench: \url{https://huggingface.co/datasets/HaochenWang/TreeBench}}

\subsection{Mathematical and Logical Reasoning}
\label{app:benchmarks-reasoning}

\textbf{MathVista-Mini}~\citep{lu2024mathvista} is the testmini split of MathVista, covering mathematical reasoning over diagrams, charts, and natural images.\footnote{MathVista-Mini: \url{https://huggingface.co/datasets/AI4Math/MathVista}}

\textbf{MathVerse}~\citep{zhang2025mathverse} evaluates diagram understanding in visual mathematics.
We use its vision-only version, which presents the problem information entirely in the image.\footnote{MathVerse: \url{https://huggingface.co/datasets/AI4Math/MathVerse}}

\textbf{VisuLogic}~\citep{xu2026visulogic} tests nonverbal logical reasoning through visual patterns involving quantity, space, and attributes.\footnote{VisuLogic: \url{https://huggingface.co/datasets/VisuLogic/VisuLogic}}

\subsection{General Multimodal Understanding}
\label{app:benchmarks-general}

\textbf{HallusionBench}~\citep{guan2024hallusionbench} diagnoses language hallucinations and visual illusions through questions that require image context.\footnote{HallusionBench: \url{https://huggingface.co/datasets/rayguan/HallusionBench}}

\textbf{ChartQA-Pro}~\citep{masry2025chartqapro} covers diverse chart types and question formats, including conversational, hypothetical, and unanswerable questions.\footnote{ChartQA-Pro: \url{https://huggingface.co/datasets/ahmed-masry/ChartQAPro}}

\textbf{InfographicVQA}~\citep{mathew2022infographicvqa} requires joint reasoning over text, layout, graphics, and data visualizations to answer questions about infographics.\footnote{InfographicVQA: \url{https://site.docvqa.org/datasets/infographicvqa}}